\documentclass[10pt,journal, compsoc]{IEEEtran}
\usepackage{epsfig}
\usepackage{color}
\usepackage{graphicx}
\usepackage{epstopdf}
\usepackage{amsmath}
\usepackage{amssymb}
\usepackage{algorithm}
\usepackage[noend]{algorithmic}
\usepackage{url}
\usepackage{dsfont}
\usepackage{epstopdf}
\usepackage{subfigure}
\usepackage{mathtools}
\usepackage{multirow}
\usepackage{footnote}
\usepackage{booktabs}
\definecolor{DeltaColor}{rgb}{0.039,0.73,0.71}
\definecolor{SigmaColor}{rgb}{0.98,0.45,0.0}
\definecolor{AlphaColor}{rgb}{0,0,0.8}
\definecolor{BetaColor}{rgb}{0.8,0,0.8}
\definecolor{GammaColor}{rgb}{0.514,0.34,0.224}
\definecolor{EpsilonColor}{rgb}{0.353,0.725,0.906}
\definecolor{PurpleColor}{rgb}{0.5,0,0.7}
\definecolor{OrangeColor}{rgb}{1,0.5,0}

\newcommand\red[1]{{{#1}}}
\newcommand\redred[1]{{{#1}}}
\ifCLASSOPTIONcompsoc
  \usepackage[nocompress]{cite}
\else
  \usepackage{cite}
\fi

\begin{document}
%
\title{AlphaPose: Whole-Body Regional Multi-Person Pose Estimation and Tracking in Real-Time}

\author{Hao-Shu Fang$^{*}$,
        Jiefeng Li$^{*}$,
        Hongyang Tang,
        Chao Xu$^{\$}$,
        Haoyi Zhu$^{\$}$,
        Yuliang Xiu,
        Yong-Lu Li,
        and~Cewu Lu$^{\dagger}$,~\IEEEmembership{Member,~IEEE}
\IEEEcompsocitemizethanks{\IEEEcompsocthanksitem Hao-Shu Fang, Jiefeng Li, Hongyang Tang, Chao Xu, Haoyi Zhu, Yong-Lu Li and Cewu Lu are with the Department
of Electrical and Computer Engineering, Shanghai Jiao Tong University, Shanghai, 200240, China.
\IEEEcompsocthanksitem * denotes the first two authors contribute equally to the manuscript, email: fhaoshu@gmail.com, ljf\_likit@sjtu.edu.cn. \$ denotes the fourth and fifth author contribute equally.
\IEEEcompsocthanksitem Corresponding author: Cewu Lu, email: lucewu@sjtu.edu.cn}
}

%
%

\markboth{IEEE TRANSACTIONS ON PATTERN ANALYSIS AND MACHINE INTELLIGENCE, VOL. XX, NO. XX, XXX. XXXX}%
{****for peer review only****}
%



\IEEEtitleabstractindextext{%
\begin{abstract}
Accurate whole-body multi-person pose estimation and tracking is an important yet challenging topic in computer vision. To capture the subtle actions of humans for complex behavior analysis, whole-body pose estimation including the face, body, hand and foot is essential over conventional body-only pose estimation. In this paper, we present AlphaPose, a system that can perform accurate whole-body pose estimation and tracking jointly while running in realtime. To this end, we propose several new techniques: Symmetric Integral Keypoint Regression (SIKR) for fast and fine localization, Parametric Pose Non-Maximum-Suppression (P-NMS) for eliminating redundant human detections and Pose Aware Identity Embedding for jointly pose estimation and tracking. During training, we resort to Part-Guided Proposal Generator (PGPG) and multi-domain knowledge distillation to further improve the accuracy.
Our method is able to localize whole-body keypoints accurately and tracks humans simultaneously given inaccurate bounding boxes and redundant detections. We show a significant improvement over current state-of-the-art methods in both speed and accuracy on COCO-wholebody, COCO, PoseTrack, and our proposed Halpe-FullBody pose estimation dataset. Our model, source codes and dataset are made \textbf{publicly available at https://github.com/MVIG-SJTU/AlphaPose}.\footnote[1].
\end{abstract}


}

\maketitle

\IEEEdisplaynontitleabstractindextext

\ifCLASSOPTIONpeerreview
\begin{center} \bfseries EDICS Category: 3-BBND \end{center}
\fi
%
\IEEEpeerreviewmaketitle

\IEEEraisesectionheading{\section{Introduction}\label{sec:introduction}}

%
%
%
%

\IEEEPARstart{F}{ull} body human pose estimation is a fundamental challenge for computer vision. It has many applications in human-computer interaction~\cite{wang2018human}, film industry~\cite{moeslund2006survey}, action recognition~\cite{pang2018human}, etc.

In this work, we focus on the problem of multi-person full body pose estimation. In conventional body-only pose estimation, recognizing the pose of multiple persons in the wild is more challenging than recognizing the pose of a single person in an image~\cite{sapp2010cascaded,sun2012conditional,ladicky2013human,newell2016stacked,wei2016convolutional}. Previous attempts approached this problem by using either a top-down framework~\cite{pishchulin2012articulated,gkioxari2014using}  or
a bottom-up framework ~\cite{chen2015parsing,pishchulin16cvpr,insafutdinov16ariv}.

Our approach follows the top-down framework, which first detects human bounding boxes and then estimates the pose within each box independently. \red{For top-down based methods, although their performances are dominant on common benchmarks~\cite{MSCOCO,andriluka14cvpr}, such methodology has some drawbacks.} Since the detection stage and the pose estimation stage are separated, i) if the detector fails, there is no cue for the pose estimator to \redred{recover the human pose}, and ii) current researchers adopt strong human detectors for accuracy, which makes the two step processing slow in inference. To solve these drawbacks of the top-down framework, we propose a new methodology to make it efficient and reliable in practice. To alleviate the missing detection problem, we lower the detection confidence and NMS threshold to provide more candidates for subsequent pose estimation. The resulted redundant poses from redundant boxes are then eliminated by a parametric pose NMS, which introduces a novel pose distance metric to compare pose similarity. A data-driven approach is applied to optimize the pose distance parameters. We show that with such strategy, a top-down framework with YOLOV3-SPP detector can achieve on par performance with the state-of-the-art detectors while achieving much higher efficiency. Furthermore, to speed up the top-down framework during inference, we design a multi-stage concurrent pipeline in AlphaPose, which allows our framework to run in realtime.

\begin{figure}[t]
\begin{center}
   \includegraphics[width=0.95\linewidth]{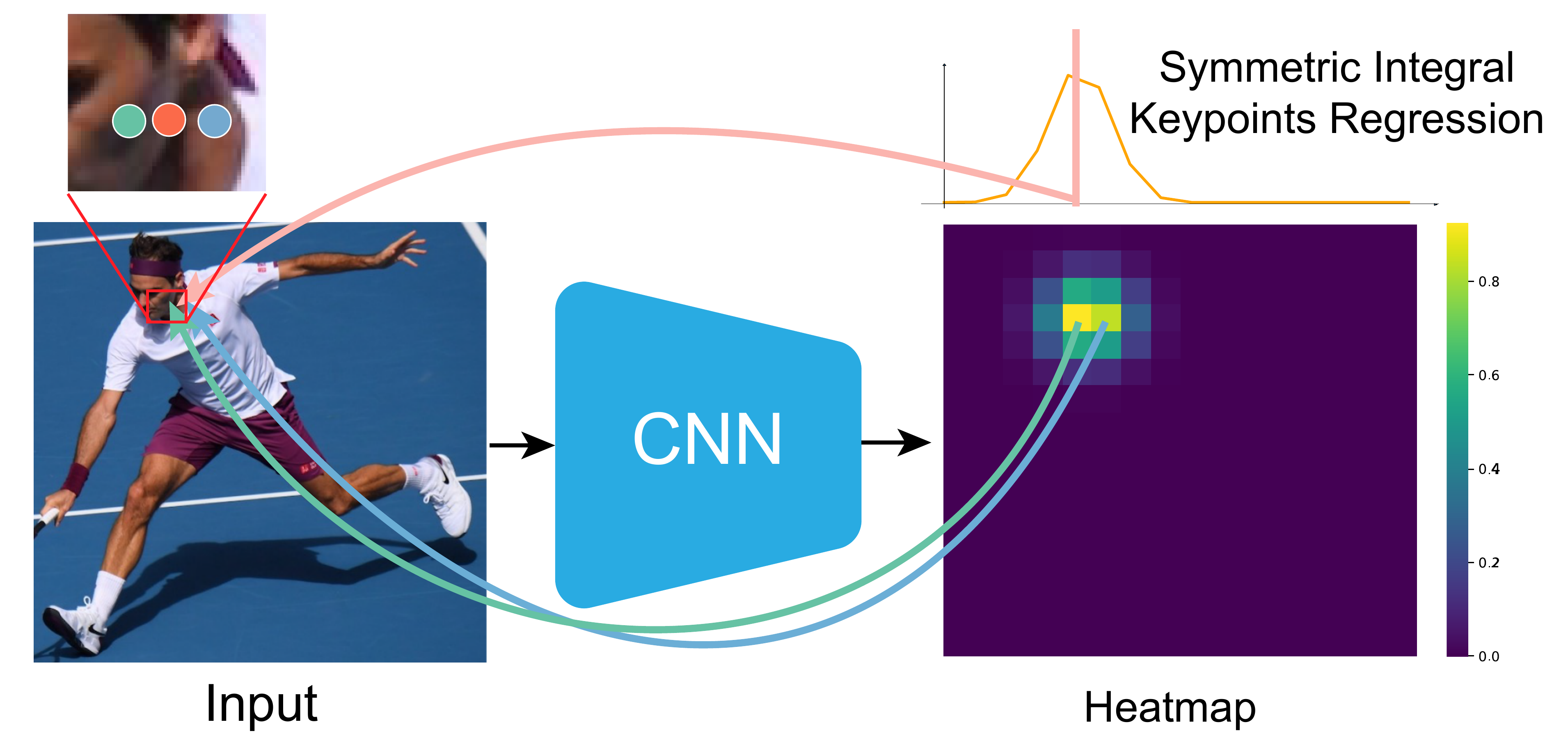}
\end{center}
   \caption{The quantization error caused by heatmap (green and blue lines). With our symmetric integral keypoints regression (pink line), we can resolve the localization error.}
\label{fig:integral}
\end{figure}

Beyond body-only pose estimation, \emph{full body} pose estimation in the wild is more challenging as it faces several extra problems. For both top-down framework and bottom-up framework, the currently most used representation for keypoint is the heatmap~\cite{tompson2014joint}. And the heatmap size is usually the quarter of the input image due to the limit of computation resources. However, for localizing the keypoints of body, face and hands simultaneously, such representation is unsuitable since it is incapable of handling the large scale variation across different body parts. A major problem is referred as the quantization error. As illustrated in Fig.~\ref{fig:integral}, since the heatmap representation is discrete, both the adjacent grids on heatmap may miss the correct position. This is not a problem for body pose estimation since the correct area are usually large. However, for fine-level keypoints on hands and face, it is easy to miss the correct position.
 
To solve this problem, previous methods either adopt additional sub-networks for hand and face estimation~\cite{cao2018openpose}, or adopt ROI-Align to enlarge the feature map~\cite{jin2020whole}. However, both methods are computation expensive, especially in multi-person scenario. In this paper, we propose a novel symmetric integral keypoints regression method that can localize keypoints in different scales accurately. It is the first regression method that can have the accuracy on par with heatmap representation while eliminate the quantization error.

Another problem for the full body pose estimation is the lack of training data. Unlike the frequent studied body pose estimation with abundant datasets~\cite{MSCOCO,li2019crowdpose}, there is only one dataset~\cite{jin2020whole} for the full body pose estimation. To promote development in this area, we annotate a new dataset named Halpe for this task, which includes extra essential joints not available in~\cite{jin2020whole}.
To further improve the generality of top-down framework for full body pose estimation in the wild, two key components are introduced.  We adopt a Multi-Domain Knowledge Distillation to incorporate training data from separate body part datasets. To alleviate the domain gap between different datasets and the imperfect detection problem, we propose a novel part-guided human proposal generator (PGPG) to augment training samples.
By learning the output distribution of a human detector for different poses, we can simulate the generation of human bounding boxes, producing a large sample of training data.

At last, we introduce a pose-aware identity embedding to enable simultaneous human pose tracking within our top-down framework. A person re-id branch is attached on the pose estimator and we perform jointly pose estimation and human identification. With the aid of pose-guided region attention, our pose estimator is able to identify human accurately. Such design allows us to achieve realtime pose estimation and tracking in an unified manner.

This manuscript extends our preliminary work published at the conference ICCV 2017~\cite{fang2017rmpe} along the following aspects:
\begin{itemize}
    \item We extend our framework to full body pose estimation scenario and propose a new symmetric integral keypoint localization network for fine-level localization.
    \item We extend our pose guided proposal generator to incorporate with multi-domain knowledge distillation on different body part dataset.
    \item We annotate a new whole-body pose estimation benchmark (136 points for each person) and make comparisons with previous methods.
    \item We propose the pose-aware identity embedding that enable pose tracking in our top-down framework in a unified manner.
    \item This work documents the release of AlphaPose, which achieves both accurate and realtime performance. Our library facilitated many researchers and has been starred for over 6,000 times on GitHub.
\end{itemize}

\section{Related Work}
\red{In this section, we first briefly review papers in multi-person pose estimation, which provides background knowledge of human pose estimation. In Sec.~\ref{sec:review_whole} we review related works in multi-person \emph{whole-body} pose estimation and discuss a key issue in the current literature. In Sec.~\ref{sec:review_ik} we review integral regression based keypoint localization and clarify our improvements toward previous works. In Sec.~\ref{sec:review_track} we review pose tracking and summarize the connection and differences between previous works and our method.}
\begin{figure*}[hbt]
\centering
\begin{tabular}{@{\hspace{0mm}}c@{\hspace{1mm}}c@{\hspace{1mm}}c@{\hspace{1mm}}c@{\hspace{1mm}}c}
\includegraphics[width=0.95\linewidth]{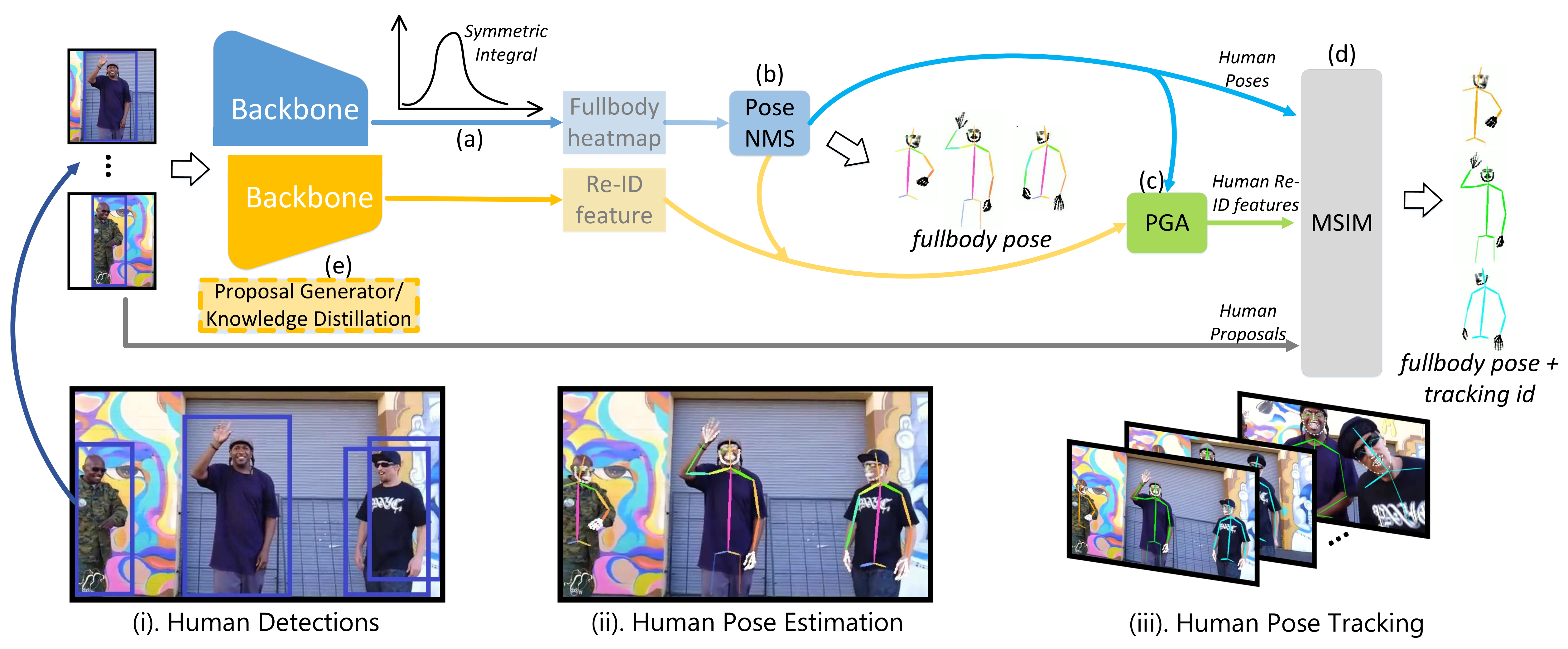}
\end{tabular}
\caption{Illustration of our full-body pose estimation and tracking framework. Given an input image, we first obtain (i) human detections using off-the-shelf object detectors like YoloV3 or EfficientDet. For each detected human, we crop and resize it and forward it through pose estimation and tracking networks to obtain full body human pose and Re-ID features. The backbone of these two networks can either be separated for adaptation to different pose configurations, or share the same weights for fast inference (thus misaligned in the figure). The (a) symmetric integral regression is adopted for fine-level keypoint localization. We adopt (b) pose NMS to eliminate redundant poses. The (c) pose-guided alignment (PGA) module is applied on the predicted human re-id feature to obtain pose-aligned human re-id features. The (d) multi-stage identity matching (MSIM) utilize the human poses, re-id features and detected boxes to produce the final tracking identity. During training, proposal generator and knowledge distillation (e) is adopted to improve the generalization ability of the networks.}
\vspace{-0.1in}
\label{fig:framework}
\end{figure*}

\subsection{Multi Person Pose Estimation}
\noindent{\bf Bottom-up Approaches}
Bottom-up approaches are also called part-based approaches in the early. These approaches firstly detect all possible body parts in an image and then group them into individual skeletons. Representative works \cite{chen2015parsing,gkioxari2014using,Iqbal_ECCVw2016,pishchulin16cvpr,insafutdinov16ariv,kreiss2019pifpaf,kreiss2021openpifpaf} are reviewed. Chen \emph{et al.}~\cite{chen2015parsing} present an approach to parse largely occluded people by graphical model which models humans as flexible compositions of body parts. Gkiox \emph{et al}~\cite{gkioxari2014using} use k-poselets to jointly detect people and predict locations of human poses by a weighted average of all activated poselets. Pishchulin \emph{et al}~\cite{pishchulin16cvpr} propose DeepCut to first detect all body parts, and then label, filter and assemble these parts via integral linear programming. A stronger part detector based on ResNet\cite{he2016deep} and a better incremental optimization strategy is proposed by Insafutdinov \emph{et al} \cite{insafutdinov16ariv}, named DeeperCut. Openpose~\cite{cao2017realtime,cao2018openpose} introduces Part Affinity Fields (PAFs) to encode association scores between body parts with individuals and solves the matching problem  by decomposing it into a set of bipartite matching subproblems. Newell \emph{et al}\cite{newell2017associative} learn an identification tag for each part detected to indicate which individuals it belongs to, named associative embedding. Cheng \emph{et al}\cite{bowen2020higherhret} use a powerful multi-resolution network\cite{sun2019deep} as backbone and high-resolution feature pyramids to learn scale-aware representations. \red{OpenPifpaf~\cite{kreiss2019pifpaf,kreiss2021openpifpaf} proposes a Part Intensity Field (PIF) and a
Part Association Field (PAF) to localize and associate body parts respectively.}

While bottom-up approaches have demonstrated good performance, their body-part detectors can be vulnerable since only small local regions are considered and face the scale variation challenge when there are small persons in the image.

\vspace{2mm}
\noindent{\bf Top-down Approaches}
Our work follows the top-down paradigm like others\cite{pishchulin2012articulated,fang2017rmpe,he2017mask,chen2018cascaded,xiao2018simple,sun2019deep}, which firstly obtains the bounding box for each human body through object detector and then performs single-person pose estimation on cropped image. Fang \emph{et al}\cite{fang2017rmpe} propose symmetric spatial transformer network to solve the problem on imperfect bounding boxes with huge noise given by human body detector. Mask R-CNN\cite{he2017mask} extends Faster R-CNN\cite{ren2015faster} by adding a pose estimation branch in parallel with existing bounding box recognition branch after ROIAlign, enabling end-to-end training. \red{PandaNet~\cite{benzine2020pandanet} proposes an anchor based method to predict multi-person 3D pose estimation in a single shot manner and achieved high efficiency.} Chen \emph{et al}\cite{chen2018cascaded} use a feature pyramid network to localize simple joints and a refining network which integrates features of all levels from previous network to handle hard joints. A simple-structured network \cite{xiao2018simple} with ResNet\cite{he2016deep} as backbone and a few deconvolutional layers as up-sampling head shows effective and competitive results. Sun \emph{et al}\cite{sun2019deep} present a powerful high-resolution network, where a high-resolution subnetwork is established in the first stage, and high-to-low resolution subnetworks are added one by one in parallel in subsequent stages, conducting repeated multi-scale feature fusions. \red{Bertasius \emph{et al}\cite{bertasius2019learning} extend from images to videos and propose a method for learning pose warping on sparsely labeled videos.}

Although state-of-the-art top-down approaches achieve remarkable precision on popular large-scale benchmark, \red{The two step paradigm makes them slow in inference compared with the bottom-up approaches}. In addition, the lack of library-level framework implementation hinders them from being applied to the industry. \red{Thus we present AlphaPose in this paper, in which we develop a multi-stage pipeline to simultaneously process the time-consuming steps and enable fast inference.}

\vspace{2mm}
\noindent{\bf One-stage Approaches}
Some approaches need neither post joints grouping nor human body bounding boxes detected in advance. They locate human bodys and detect their own joints simultaneously to improve low efficiency in two-stage approaches. Representative works include CenterNet~\cite{xingyi2019object}, SPM~\cite{nie2019single}, DirectPose~\cite{tian2019directpose}, and Point-set Anchor~\cite{wei2020point}. However, these approaches do not achieve high precision as top-down approaches, partly because body center map and dense joint displacement maps are high-semantic nonlinear representations and make it difficult for the networks to learn. 

\subsection{Whole-Body Keypoint Localization}
\label{sec:review_whole}
Unified detection of body, face, hand and foot keypoints for multi-person is a relative new research topic and few methods have been proposed. OpenPose~\cite{cao2018openpose} developed a cascaded method. It first detects body keypoints using PAFs~\cite{cao2017realtime} and then adopts two separate networks to estimate face landmarks and hand keypoints. Such design makes it time inefficient and consumes extra computation resources. Hidalgo \emph{et al}.~\cite{hidalgo2019single} propose a single network to estimate the whole body keypoints. However, due to its one-step mechanism, the output resolution is limited and thus decrease its performance on fine-level keypoints such as faces and hands. \red{Jin \emph{et al}.~\cite{jin2020whole} propose a ZoomNet that used ROIAlign to crop the hand and face region on the feature maps and predict keypoints on the resized feature maps. All these methods adopt the heatmap representation for keypoint localization due to its dominant performance on body keypoints. However, the mentioned quantization problem of heatmap would decrease the accuracy of face and hand keypoints. The requirement of large-size input also consumes more computation resources. In this paper, we argue that soft-argmax presentation is more suitable for whole-body pose estimation and proposed an improved version of soft-argmax that yields higher accuracy.}  Jin \emph{et al}.~\cite{jin2020whole} also extended the COCO dataset to whole-body scenario. However, some joints like head and neck are not presented in this dataset, which is essential in tasks like mesh reconstruction. Meanwhile, the face annotation is incompatible with that in 300LW. In this paper, we contribute a new in-the-wild multi-person whole-body pose estimation benchmark. We annotate 40K images from HICO-DET~\cite{chao2018learning} \red{with Creative Common license}\footnote{\url{https://creativecommons.org/licenses/}} as the training set and extend the COCO keypoints validation set (6K instances) as our test set. Experiments on this benchmark and COCO-Wholebody demonstrate the superiority of our method.
\red{
\subsection{Integral Keypoints Localization}
\label{sec:review_ik}
Heatmap is a dominant representation for joint localization in the field of human pose estimation. The read-out locations from heatmaps are discrete numbers since heatmaps only describe the likelihood of joints occurring in each spatial grid, which leads to inevitable quantization error. As denoted in Sec.~\ref{sec:review_whole}, we argue that \textit{soft-argmax} based integral regression is more suitable for whole-body keypoints localization. Several previous works have studied the soft-argmax operation to read continuous joint locations from heatmaps~\cite{finn2015learning,yi2016lift,luvizon20182d,nibali20193d,sun2018integral,luvizon2019human}. Specifically, Luvizon \emph{et al.}~\cite{luvizon20182d,luvizon2019human} and Sun \emph{et al.}~\cite{sun2018integral} apply the \textit{soft-argmax} operation to single-person 2D/3D pose estimation successfully. However, two drawbacks exist in these works that decrease their accuracy in pose estimation. We summarize the drawbacks as asymmetric gradient problem and size-dependent keypoint scoring problem. Details of these problems are provided in Sec.~\ref{sec:SIKR}, as well as our proposed new gradient design and keypoint scoring method. By solving these problems, we provide a new keypoint regression method with higher accuracy, and it shows good performance in both whole-body pose estimation and body-only pose estimation.}

\subsection{Multi Person Pose Tracking}
\label{sec:review_track}
Multi person pose tracking is extended from multi person pose estimation in videos, which gives each predicted keypoint the corresponding identity over time. \redred{Similar to pose estimation literature,} it can be divided into two categories: top-down\cite{xiao2018simple,xiu2018pose,ning2019lighttrack,girdhar2018detect,wang2020combining, guo2018multi, bao2020pose, yang2021learning, snower202015} and bottom-up\cite{raaj2019efficient,jin2019multi,kreiss2021openpifpaf}. Based on the bottom-up pose estimation methods, \cite{raaj2019efficient,jin2019multi} use the detected keypoints to build temporal and spatial graphs which aims to link corresponding individual body by solving an optimization problem. 
However, the prerequisite of temporal and spatial graphs prevent graph-cut optimization from running in online manner, which makes them quite time-consuming and memory-inefficient. \cite{girdhar2018detect} utilizes a 3D MaskRCNN to estimate person tubes and poses simultaneously. \cite{wang2020combining} proposes forward and backward bounding box propagation strategy to eliminate the issue of missed detection. The input of these methods is a whole video sequence, which cannot achieve online tracking.  Some other top-down methods allow to input a single frame, and then use designed poseflow\cite{xiu2018pose}, \red{GCN\cite{ning2019lighttrack,bao2020pose}}, optical flow\cite{xiao2018simple} \red{or transformer~\cite{snower202015}} for identity matching. \red{Yang \emph{et al}~\cite{yang2021learning} predict current poses given historical pose sequences and merge them with the pose estimation results from the current frame. A drawback of these methods is that they rely on the spatial continuity of the poses only, which may not be satisfied when the online image stream is unstable or humans are moving rapidly. Specifically,} \cite{zhang2019fastpose} proposes to use \redred{re-ID} feature to tackle tracking problem. \red{Our tracking method also explicitly adopts the human \redred{re-ID} features to solve this problem. Compared with \cite{zhang2019fastpose}, we design a pose-guided \redred{re-ID} feature extraction to avoid potential background noise. Moreover, we design a multi-stage information merging method to utilize the boxes, poses, and \redred{re-ID} features simultaneously.}

\section{Whole-Body Multi Person Pose Estimation}
The whole pipeline of our proposed method is illustrated in Figure \ref{fig:framework}. In this section, we introduce the details of our pose estimation, which is shown in the top row of Fig.~\ref{fig:framework}.

\subsection{Symmetric Integral Keypoints Regression}
\label{sec:SIKR}
\red{As mentioned in Sec.~\ref{sec:review_ik}, there exist two problems in conventional soft-argmax operation for keypoint regression. We illustrate them in the following sections and propose our novel solution.}

\subsubsection{\red{Asymmetric gradient problem}} The \textit{soft-argmax} operation, \redred{also known as} \textit{integral regression}, is differentiable, which turns heatmap based approaches into regression based approaches and allows end-to-end training. The \textit{integral regression} operation is defined as:
\begin{equation}
    \hat{\mu} = \sum x \cdot p_x,
\end{equation}
where $x$ is coordinate of each pixel and $p_x$ denotes the pixel likelihood on heatmap \red{after normalization}. During training, the loss function is applied to minimize the $\ell_1$ norm between the predicted joint locations $\hat{\mu}$ and ground-truth locations $\mu$: $\mathcal{L}_{reg} = \|\mu - \hat{\mu}\|_1$. The gradient of each pixel can be formulated as:
\begin{equation}
    \frac{\partial \mathcal{L}_{reg}}{\partial p_x} = x \cdot \text{sgn}(\hat{\mu} - \mu).
\end{equation}
\red{Notice that the gradient amplitude is asymmetric. The absolute value of the gradient is determined by the absolute position (\emph{i.e.} $x$) of the pixel instead of the relative position to the ground truth. It denotes that given the same distance error, the gradient becomes different when the keypoint locates at a different position. This asymmetry breaks the translation invariance of the CNN network, which leads to performance degradation.}

\noindent{\bf Amplitude Symmetric Gradient} To improve the learning efficiency, we propose an amplitude symmetric gradient (ASG) function \red{in backward propagation}\redred{, which is an approximation to the true gradient}:
\begin{equation}
   \redred{\delta_{ASG}} = A_{grad} \cdot \text{sgn}(x - \hat{\mu}) \cdot \text{sgn}(\hat{\mu} - \mu),
\end{equation}
where $A_{grad}$ denotes the amplitude of gradients. \red{It is a constant that we manually set as 1/8 of the heatmap size and we give the derivation in the next paragraph.} Using our symmetric gradient, the gradient distribution is centred at the predicted joint locations $\hat{\mu}$. In the learning process, this symmetric gradient distribution can better utilize the advantage of heatmaps and approximate the ground-truth locations in a more direct manner. For example, assume the predicted location $\hat{\mu}$ is higher than the ground truth $\mu$.
On one hand, the network tends to suppress the heatmap values on the right side of $\hat{\mu}$, because they have positive gradients; on the other hand, the heatmap values on the left side of $\hat{\mu}$ will be activated because of their negative gradients.

\noindent{\bf Stable gradients for ASG}

\red{Here, we conduct a Lipschitz analysis to derive the value of $A_{grad}$ and show that ASG can provide a more stable gradient for training.}
Recall that $f$ denotes the objective function that we want to minimize. We say that $f$ is $L$-smooth if:
\begin{equation}
    \| \nabla_{\theta}f(\theta + \Delta\theta) - \nabla_{\theta}(\theta)  \| \leq L\| \Delta\theta \|,
\end{equation}
where $\theta$ is the network parameters and $\nabla$ denotes the gradient. The objective function can be re-written as:
\begin{equation}
    \nabla_{\theta}f = \nabla_{\theta}\mathcal{L}(\mu, h(z)) = \nabla_{z}\mathcal{L}(\mu, h(z))\nabla_{\theta}z,
\end{equation}
where $z$ denotes the logits that are predicted by the network, and $\hat{\mu} = h(z)$ denotes the \red{composition of the normalization and soft-argmax functions}. Here, we assume the gradient of the network is smooth and only analyze the \red{composition function}, {i.e.}:
\begin{equation}
    \| \nabla_{z}\mathcal{L}(\mu, h(z + \Delta z)) - \nabla_{z}\mathcal{L}(\mu, h(z)) \|.
    \label{eq:lip}
\end{equation}
In the conventional integral regression, we have:
\begin{equation}
    \nabla_{z}\mathcal{L}(\mu, h(z)) = (x - \hat{\mu})\cdot p_x.
\end{equation}
In this case, Eq.~\ref{eq:lip} is equivalent to:
\begin{equation}
    \| (x - \hat{\mu} - \Delta\hat{\mu})(p_x + \Delta p_x) - (x - \hat{\mu})\cdot p_x \|.
    \label{eq:lip-integral}
\end{equation}
Note that $x$ can be a arbitrary position on the heatmap. Denoting the heatmap size as $W$, we have $\| x - \hat{\mu} \| \leq W$ over the whole dataset. Therefore, we derive the Lipschitz constant of integral regression as:
\begin{equation}
\begin{aligned}
    &\| \nabla_{z}\mathcal{L}(\mu, h(z + \Delta z)) - \nabla_{z}\mathcal{L}(\mu, h(z)) \| \\
    \leq &\| W(p_x + \Delta p_x) - W p_x \| = W \| \Delta p_x \| \\
    = &W \cdot L_s \cdot \| \Delta z\|,
\end{aligned}
\end{equation}
where $L_s$ is the Lipschitz constant of the \red{normalization function~\cite{gao2017properties,gouk2021regularisation}}. It shows that the \red{conventional integral regression} multiplies a factor $W$ to the Lipschitz constant of normalization.

Similarly, we can derive the Lipschitz constant of the proposed amplitude symmetric function. Firstly, the gradient of the logits is:
\begin{equation}
\begin{aligned}
   | \nabla_{z}\mathcal{L}(\mu, h(z)) | &= | A_{grad}\cdot p_x \cdot (1 + \sum_{x_i < \hat{\mu}} p_{x_i} - \sum_{x_i > \hat{\mu}} p_{x_i}) | \\
   &\leq 2 \cdot A_{grad} \cdot p_x.
\end{aligned}
\end{equation}
We set $A_{grad} = W / 8 $ to make the average norm of the gradient the same as integral regression. Specifically,
\begin{equation}
    E_{x}[|(x - \hat{\mu})p_x|] = E_{x}[|x - \hat{\mu}|]p_x = \frac{W}{4} \cdot p_x.
\end{equation}
The Lipschitz constant of the proposed amplitude symmetric function derived as:
\begin{equation}
\begin{aligned}
    &\| \nabla_{z}\mathcal{L}(\mu, h(z + \Delta z)) - \nabla_{z}\mathcal{L}(\mu, h(z)) \| \\
    \leq &\| 2A_{grad}(p_x + \Delta p_x) - 2A_{grad} p_x \| = \frac{W}{4} \| \Delta p_x \| \\
    = &\frac{W}{4} \cdot L_s \cdot \| \Delta z\|.
\end{aligned}
\end{equation}
It shows that the Lipschitz constant of the proposed method is $4$-times smaller than the original integral regression \red{when $A_{grad} = W / 8 $}, which indicates that the gradient space is more smooth and the model can be optimized more easily.

\subsubsection{\red{Size-dependent Keypoint Scoring Problem}}
\red{Before conducting soft-argmax}, the element-sum of the predicted heatmaps should be normalized to one, \red{i.e., $\sum p_x = 1$}.
Prior works~\cite{luvizon2019human,sun2018integral} adopt \textit{soft-max} operation, which works well in single-person pose estimation but \redred{remains a large performance gap with the state-of-the-arts} in multi-person pose estimation\redred{~\cite{sun2018integral,xiao2018simple,li2021human}}. This is because in multi-person cases, we need not only the joint locations, but also the joint confidence \redred{for pose NMS and calculating the mAP}. \redred{In previous methods}, the maximum value of the heatmap is taken as the joint confidence, \redred{which is size-dependent and not accurate.}

\red{If we adopt the one-step normalization such as \textit{soft-max}}, the maximum value of the heatmap is inversely proportional to \red{the scale of the distribution}, which highly depends on the projected size of the body joint. Therefore, a large-size joint (e.g. left-hip) will generate a smaller confidence value than a small-size joint (e.g. nose), which harms the reliability of the predicted confidence values. 

\noindent{\red{\bf{Two-step Heatmap Normalization}}} To decouple confidence prediction and integral regression, we propose a two-step heatmap normalization manner. In the first step, we perform element-wise normalization to generate the confidence heatmap $\mathbf{C}$:
\begin{equation}
    c_x = \text{sigmoid}(z_x),
\end{equation}
where $z_x$ denotes the un-normalized logits value of location $x$, $c_x$ denotes the confidence heatmap value of location $x$. Hence, the joint confidence can be indicated by the maximum value of the heatmap:
\begin{equation}
    \mathit{conf} = \max{(\mathbf{C})}.
\end{equation}
\redred{Since we use an element-wise operation \textit{sigmoid} for the first step of normalization and don't force the sum of $\mathbf{C}$ to be one, the maximum value of $\mathbf{C}$ won't be affected by the size of the joint.}
In this way, the predicted joint confidence is only related to the predicted location. In the second step, we perform global normalization to generate the probability heatmap $\mathbf{P}$:
\begin{equation}
    p_x = \frac{c_x}{\sum \mathbf{C}}.
\end{equation}
The element-sum of the probability heatmap $\mathbf{P}$ is one, which ensures the predicted joint location $\hat{\mu}$ is within the heatmap boundary and stabilizes the training process. 

\redred{To sum up, we obtain the joint confidence through the first step and obtain the joint location on the heatmap generated by the second step. An ablation study is carried out in Sec.~\ref{sec:ablation} to show the effectiveness of our normalization method.}

\subsection{Multi-Domain Knowledge Distillation}
Beyond our novel symmetric integral regression, the performance of network can further benefit from extra training data. Except for annotating a new dataset (detailed in Sec~\ref{sec:datasets}), we also adopt multi-domain knowledge distillation to train our network.  Three additional datasets are adopted, namely 300Wface~\cite{sagonas2016300}, FreiHand~\cite{zimmermann2019freihand} and InterHand~\cite{moon2020interhand2}. The details of these datasets will be introduced in Sec~\ref{sec:datasets}. Combining these datasets, our network are able to predict face and hand keypoints accurately for in-the-wild images.

During training, we construct each training batch by sampling different datasets with a fixed ratio. To be specific, 1/3 of the batch are sampled from our annotated dataset, 1/3 from the COCO-fullbody and the remaining are equally sampled from 300Wface and FreiHand. For each sample, we apply dataset-specific augmentation, which is introduced in the next section.

Although these domain specific datasets are able to provide accurate intermediate supervision, their data distribution are quite different from in-the-wild images. To solve this problem, we extend our pose-guided proposal generator in~\cite{fang2017rmpe} to full body scenario and conduct data augmentation in a unified manner.

\subsection{Part-Guided Proposal Generator}
For the two-stage pose estimation, the human proposals generated by the human detector usually produce a different data distribution from the ground-truth human boxes. Meanwhile, the spatial distribution of the face and hand are also different between the full-body images in the wild and the part-only images in the dataset. Without proper data augmentation during training, the pose estimator may not work properly in the testing phase for the detected human. 

To generate training samples with similar distribution to the output of the human detector, we propose our part-guided proposal generator. For different body parts with a tight surrounded bounding box, the proposal generator generates a new box that is inline with the distribution of the output of human detector.

Since we already have the ground truth bounding box for each part, we simplify this problem into modeling the distribution of the relative offset between the detected bounding box and the ground truth bounding box varies across different parts. To be more specific, there exists a distribution 
$$
P(\delta \text{x$_{min}$}, \delta \text{x$_{max}$}, \delta \text{y$_{min}$}, \delta \text{y$_{max}$} | p)
$$
where $\delta \text{x$_{min}$} / \delta \text{x$_{max}$}$ is the normalized offset between the left-est/right-est coordinate of a bounding box generated by human detector and the coordinates of the ground truth bounding box:
$$
\delta \text{x$_{min}$} = \frac{\text{x$_{min}^{detect}$-x$_{min}^{gt}$}}{\text{x$_{max}^{gt}$-x$_{min}^{gt}$}},
$$
$$
\delta \text{x$_{max}$} = \frac{\text{x$_{max}^{detect}$-x$_{max}^{gt}$}}{\text{x$_{max}^{gt}$-x$_{min}^{gt}$}},
$$
and similarly is $\delta \text{y$_{min}$}, \delta \text{y$_{max}$}$, $p$ is the ground truth part type. If we can model this distribution, we are able to generate many training samples that are  similar to human proposals generated by the human detector.

\begin{figure}[bt]
\begin{center}
   \includegraphics[width=1\linewidth]{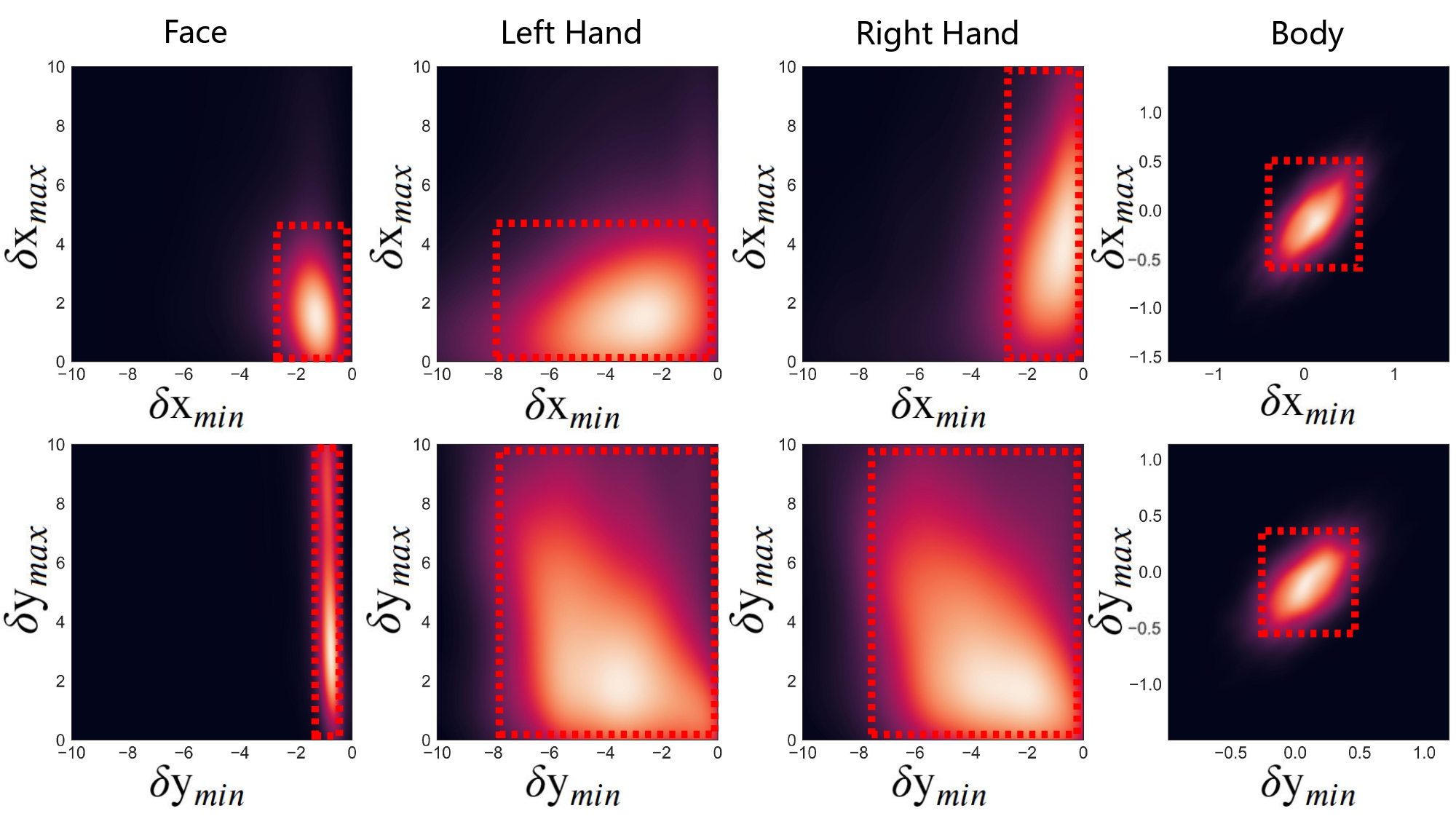}
\end{center}
   \caption{Distributions of bounding box offsets for several different body parts. \redred{The dotted boxes denote the ranges of approximated uniform distributions.} Best viewed in color.}
\vspace{-0.1in}
\label{fig:pose-distribution}
\end{figure}

To achieve that, we adopt an off-the-shelf object detector~\cite{redmon2018yolov3} and generate human detection for our Halpe-FullBody dataset. For each instances in the dataset, we separate the annotations of face, body and hand. For each separated part, we calculate the offsets between its tightly surrounded bounding box and the detected bounding box of the \textit{whole person}. Since the box variances in horizontal and vertical directions are usually independent, we simplify the the modeling of original distribution into modeling
$$
P_x(\delta \text{x$_{min}$}, \delta \text{x$_{max}$} | p),
$$
$$
P_y(\delta \text{y$_{min}$}, \delta \text{y$_{max}$} | p).
$$
After processing all the instances in Halpe-FullBody, the offsets form a frequency distribution, and we fit the data to a Gaussian mixture distribution. For different body parts, we have different Gaussian mixture parameters. We visualize the distributions and their corresponding parts in Figure~\ref{fig:pose-distribution}.

During the training phase of the pose estimator, for a training sample belonging to part $p$, we can generate additional offsets to its ground-truth bounding box by dense sampling according to $P_x(\delta \text{x$_{min}$}, \delta \text{x$_{max}$} | p)$ and $P_y(\delta \text{y$_{min}$}, \delta \text{y$_{max}$} | p)$ to produce augmented training proposals. \redred{In practice, we found that sampling in an approximated uniform distribution (the dotted red boxes in the Figure~\ref{fig:pose-distribution}) can also produce similar performance.}

\subsection{Parametric Pose NMS}
For the top-down approaches, a main drawback is the early commitment problem: if the human detector fails to detect a person, there is no recourse for the pose estimator to recover it. Most top-down based methods~\cite{papandreou2017towards, chen2018cascaded, xiao2018simple, sun2019deep} would meet this problem since they set the detection confidence to a high value to avoid redundant poses. On the contrary, we set the detection confidence to a low value (0.1 in our experiments) to ensure a high detection recall. In this case, human detectors inevitably generate redundant detections for some people, which results in redundant pose estimations. Therefore, pose non-maximum suppression (NMS) is required to eliminate the redundancies. Previous methods \cite{burgos2013merging,chen2015parsing} are either not efficient or not accurate enough. In this paper, we propose a parametric pose NMS method. Similar to the previous subsection, the pose  $P_i$, with $m$ joints is denoted as $\{\langle k_i^{1} , c_i^{1} \rangle, \ldots,\langle k_i^{m} , c_i^{m} \rangle \}$, where $k_i^j$ and $c_i^j$ are the $j^{th}$ location and confidence score of joints respectively.

\vspace{2mm}
\noindent{\bf NMS scheme} We revisit pose NMS as follows: firstly, the most confident pose is selected as reference, and some poses close to it are subject to elimination by applying \emph{elimination criterion}. This process is repeated on the remaining poses set until redundant poses are eliminated and only unique poses are reported.

\vspace{2mm}
\noindent{\bf Elimination Criterion}  We need to define pose similarity in order to eliminate the poses which are too close and too similar to each others. We define a pose distance metric $d(P_{i}, P_{j}| \Lambda)$ to measure the pose similarity, and a threshold $\eta$ as elimination criterion, where $\Lambda$ is a parameter set of function $d(\cdot)$. Our elimination criterion can be written as follows:
\begin{equation}\label{eq:elimination_criterion}
 f(\ P_{i}, P_{j}|  \Lambda,  \eta) = \mathds{1}[d(P_{i}, P_{j}|  \Lambda, \lambda) \leq \eta]
\end{equation}
If $d(\cdot)$ is smaller than $\eta$, the output of $f(\cdot)$  should be $1$, which indicates that pose $P_{i}$ should be eliminated due to redundancy with reference pose $P_{j}$.

\vspace{2mm}
\noindent{\bf Pose Distance} Now, we present the distance function $d_{pose}(P_{i}, P_{j})$. We assume that the box for $P_{i}$ is $B_{i}$. Then we define a soft matching function
\begin{eqnarray}\label{eq:count_distance}
K_{Sim}( P_{i}, P_{j}| \sigma_1 ) =  \nonumber\\
\begin{cases}
\sum_{n}  \tanh \frac{c_{i}^{n}}{\sigma_1} \cdot \tanh \frac{c_{j}^{n}}{\sigma_1}, & \text{if $k_{j}^{n}$ is within $\mathcal{B}(k_i^{n})$} \\
0 & \text{otherwise }
\end{cases}
\end{eqnarray}
where $\mathcal{B}(k_i^{n})$ is a box center at $k_i^{n}$, and each dimension of $\mathcal{B}(k_i^{n})$ is $1/10$ of the original box $B_{i}$. The $\tanh$ operation filters out poses with low-confidence scores. When two corresponding joints both have high confidence scores, the output will be close to 1. This distance softly counts the number of joints matching between poses.

The spatial distance between parts is also considered, which can be written as
\begin{equation}\label{eq:spatial_distance}
H_{Sim}( P_{i}, P_{j}| \sigma_2 ) = \sum_{n} \exp[- \frac{(k_{i}^{n}- k_{j}^{n})^2}{\sigma_2}]
\end{equation}

By combining Eqn \ref{eq:count_distance} and \ref{eq:spatial_distance}, the final distance function can be written as
\begin{equation}\label{eq:energy}
d(P_{i}, P_{j}|  \Lambda) = K_{Sim}( P_{i}, P_{j}| \sigma_1 ) +  \lambda H_{Sim}( P_{i}, P_{j}| \sigma_2 )
\end{equation}
where $\lambda$ is a weight balancing the two distances and $\Lambda = \{\sigma_1, \sigma_2, \lambda\}$. Note that the previous pose NMS \cite{chen2015parsing} set pose distance parameters and thresholds manually. In contrast, our parameters can be determined in a data-driven manner.

\vspace{2mm}
\noindent{\bf Optimization} Given the detected redundant poses, the four parameters in the eliminate criterion $f(\ P_{i}, P_{j}|\Lambda, \eta)$ are optimized to achieve the maximal mAP for the validation set. Since exhaustive search in a 4D space is intractable, we optimize two parameters at a time by fixing the other two parameters in an iterative manner. Once convergence is achieved, the parameters are fixed and will be used in the testing phase.

\begin{figure*}[hbt]
\centering
\begin{tabular}{@{\hspace{0mm}}c@{\hspace{1mm}}c@{\hspace{1mm}}c@{\hspace{1mm}}c@{\hspace{1mm}}c}
\includegraphics[width=0.95\linewidth]{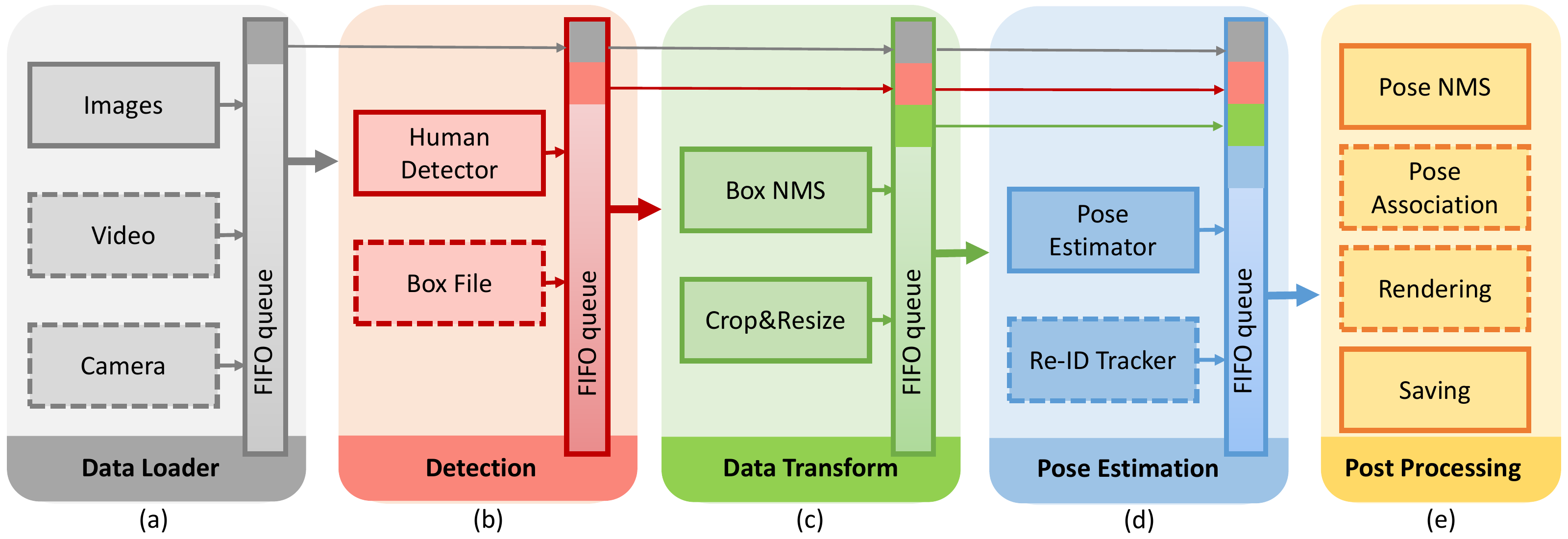}
\end{tabular}
\vspace{-0.15in}
\caption{System architecture of AlphaPose. Our system is divided into five modules, namely (a) data loading module that can take images, video or camera stream as input, (b) detection module that provides human proposals, (c) data transformation module to process the detection results and crop each single person for later modules, (d) pose estimation module that generates keypoints and /or human identity for each person, (e) post processing module that processes and saves the pose results. Our framework is flexible and each module contains several components that can be replaced and updated easily. Dashed box denotes optional components in each module. See text for more details and best viewed in color.}
\label{fig:pipeline}
\end{figure*}

\section{Multi Person Pose Tracking}
In this section, we introduce our multi person pose tracking method shown in the middle row of Fig.~\ref{fig:framework}. We attach a person \redred{re-ID} branch on the pose estimator. Thus the network can estimate human pose and \redred{re-ID} feature simultaneously. A Pose-Guided Attention Mechanism (PGA) is adopted to enhance the person identity feature. Finally, 
human proposal information (identity embedding, box and pose) are integrated by our designed Multi-Stage Identity Matching (MSIM) algorithm to achieve online realtime pose tracking.

\subsection{Pose-Guided Attention Mechanism}
Person \redred{re-ID} feature can be used to  identify the same individual from a lot of human proposals. In our top-down framework, we extract \redred{re-ID} feature from each bounding box produced by object detector. However, the quality of the \redred{re-ID} feature will be reduced by the background in the bounding box, especially when there exists other people's bodies. In order to solve this problem, we consider using the predicted human pose to construct a region where the human body is concentrated. Thus, the Pose-Guided Attention (PGA) is proposed to force the extracted features focusing on the human body of interest, and ignore the impact of the background. The insight of PGA is elaborated in ablation studies (Sec.~\ref{sec:posetrack_ablation}).

The pose estimator generates $k$ heatmaps where $k$ means the number of keypoint for each person. Then the PGA module will transform these heatmaps to an attention map ($m_A$) with a simple conv layer. Note that the $m_A$ has same size with \redred{re-ID} feature map ($m_{id}$). Therefore, we could obtain the weighted \redred{re-ID} feature map ($m_{wid}$) :
\begin{equation}\label{eq:weighted-reid-feature}
m_{wid} = m_{id}\odot m_A + m_{id}
\end{equation}
where $\odot$ means Hadamard product.

Finally, the identity embedding ($emb_{id}$) which is a 128 dimension vector is encoded by a fully-connection layer.

\subsection{ Multi-Stage Identity Matching}
For a video sequence, Let $H^i_t$ denote the i-th human proposal of t-th frame. As described above, $H^i_t$ has several features: pose ($P_t^i$), bbox ($B_t^i$) and identity embedding ($E_t^i$). 
Considering that all these features can determine the identity of a person, we design MSIM algorithm to assign the corresponding id for $H_t^i$. Assuming that the detection and tracking results of the previous t-1 frames have been obtained and stored in the tracking pool $Pl$. First, a kalman filter is used to finetune the detection features in the current frame thus make trajectories more smooth. Then we perform the first stage matching by computing the affinity matrix $M_{emb}^i$ among 
identity embedding of the \emph{t-th} frame and all embeddings existed in $Pl$. The matching rules are as follows:
\begin{eqnarray}\label{eq:emb_matching}
\begin{cases}
link(p,q), & \text{if $M_{emb}^t[p][q] = min(M_{emb}^t[p]$})\\
& \text{and $M_{emb}^t[p][q] \leq \mu_{emb}$}  \\
H_{t}^p  \quad \text{keep untracked}, & \text{otherwise }
\end{cases}
\end{eqnarray}
where $link(p,q)$ means $H_t^p$ shares the same trajectory with the q-th human proposal in $Pl$. $\mu_{emb}$ is the threshold. Here we set $\mu_{emb}$ as 0.7 following~\cite{wang2019towards}.

At the second stage, we consider both position and shape constraints for those untracked human proposals in \ref{eq:emb_matching}. Specifically, We use IOU metric between bboxes as position constraint and normalized pose distance as shape constraint. For two human proposals $H_t^i$ and $H_{t-\delta}^j$, we first resize their bbox to same scale and get the center point $c$ of each bbox. Then we compute the normalized pose 
vector by subtracting the center from each keypoint coordinates.  Finally, we can obtain the normalized pose distance ($dist_{np}$) by Eqn. \ref{eq:energy}. Therefore the fusion distance matrix of shape and location can be written as:
\begin{equation}\label{eq:fusion-distance}
M_{f}^t = (1-IOU) + \lambda_{np}\times dist_{np}
\end{equation}
Where $IOU$ and $dist_{np}$ denote the matrix formed by IOU-function and normalized pose distance among untracked proposals and $Pl$. $\lambda_{np}$ is weight to balance location and shape distance matrix.

Here we also use a threshold $\mu_{f}$ to filter unmatched proposals like Eqn \ref{eq:emb_matching} and empirically set it as 0.5.

In order to match the tracklets that are not very similar with previous frames, we appropriately lower the threshold and repeat the above stage. If there is still no matched proposal, we think that this is a newly tracklet, so a new id will be assigned to it.
\subsection{Joint Training Strategy}
In order to simplify the training process of the whole network, we simultaneously train the pose estimator and the \redred{re-ID} branch. Our network is trained on COCO\cite{MSCOCO} and PoseTrack\cite{andriluka2018posetrack} dataset. PoseTrack has both pose and identity annotation, while COCO only has pose annotation. Therefore, when training on COCO, the gradient contributed by the \redred{re-ID} branch does not participate in back propagation.  We follow the loss balanced strategy in \cite{kendall2018multi} to jointly optimize pose and identification sub-task.

\section{AlphaPose}
In this section, we present AlphaPose\footnote{Available at \url{https://github.com/MVIG-SJTU/AlphaPose}}, the first jointly whole-body pose estimation and tracking system.

\subsection{Pipeline}
A drawback of two-step framework is the limitation of the inference speed.
To facilitate the fast processing of large-scale data, we design a five-stage pipeline with multi-processing implementation to speed up our inference. Fig.~\ref{fig:pipeline} illustrates our AlphaPose pipelining mechanism. We divide the whole inference process into five modules, following the principle that each module consumes similar processing time. During inference, each module is hosted by an independent process or thread. Each process communicates with subsequent processes with a First-In-First-Out queue, that is, it stores the computed results of current module and the following modules directly fetch the results from the queue. With such design, these modules are able to run in parallel, resulting in a significant speed up and enabling real-time application.

\begin{figure}[tb]
\begin{center}
   \includegraphics[width=1\linewidth]{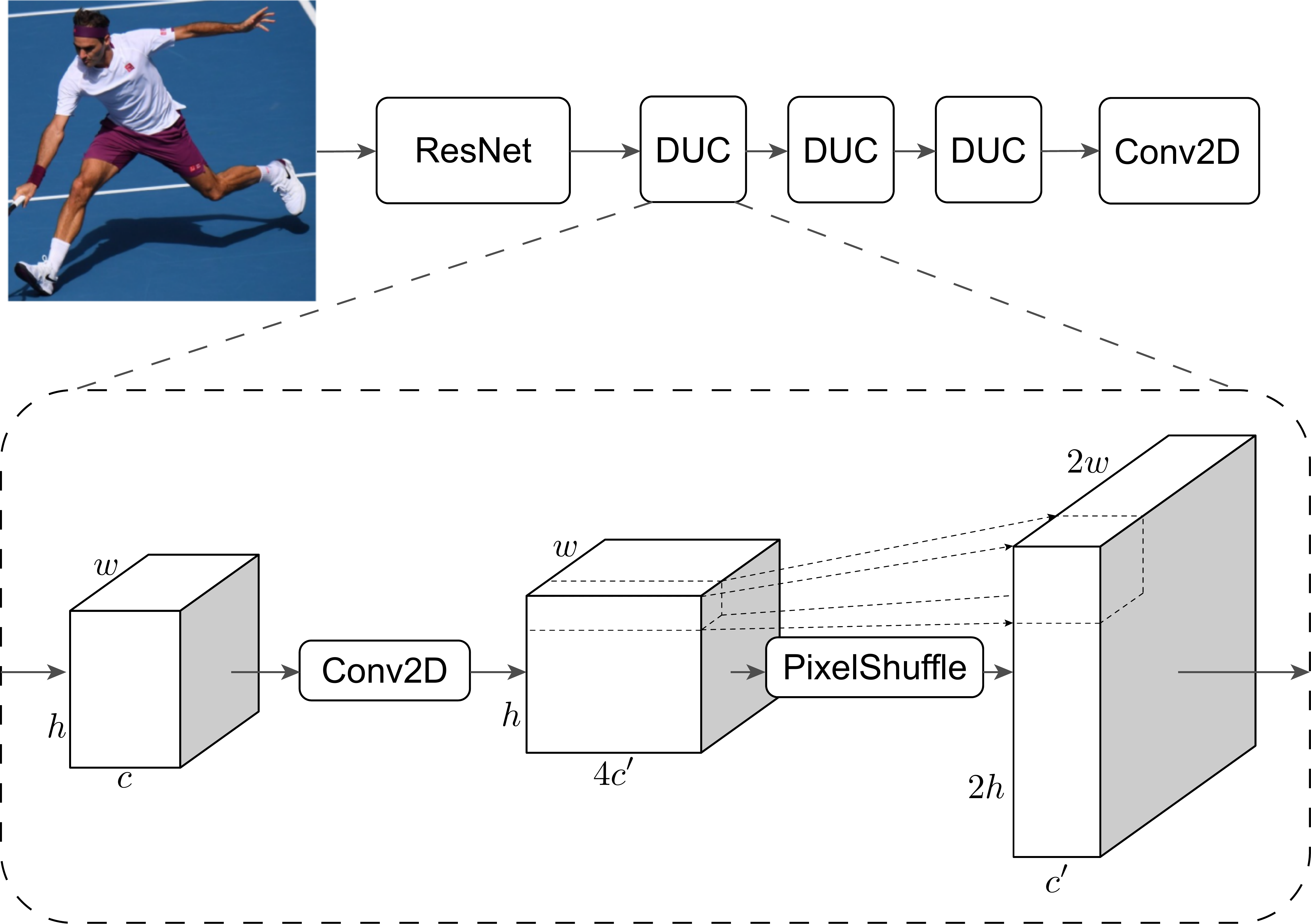}
\end{center}
\vspace{-0.1in}
   \caption{Network architecture of FastPose. Firstly, ResNet is adopted as the network backbone. Then, DUC modules are applied for up-sampling. Finally, a $1\times1$ convolution is utilized to generate heatmaps.}
\label{fig:fastpose}
\end{figure}

\begin{table}[tb]
	\begin{center}
	
    \resizebox{\linewidth}{!}{
		\begin{tabular}{c|c|c|cccc|c}
			\toprule
			DataSet  & \#Kpt & Wild &   Body   &    Hand    &   Face & HOI & Total \\ 
			&       &        &    Kpt    &    
			Kpt     &    Kpt & &Instances \\
			\midrule\midrule
			\emph{MPII}~\cite{andriluka14cvpr}      & 16 & \checkmark &\checkmark &    & & & 40K \\
			\emph{CrowdPose}~\cite{li2019crowdpose} & 14 & \checkmark  & \checkmark & & &  & 80K  \\ 
			\emph{PoseTrack}~\cite{andriluka2018posetrack}  & 15 & \checkmark  & \checkmark & & &  & 150K  \\ 
			\emph{COCO}~\cite{MSCOCO}    & 17  & \checkmark & \checkmark &    & & & 250K  \\
			\midrule
			\emph{OneHand10K}~\cite{wang2018mask}             & 21 &\checkmark  &    & \checkmark & & & 10K \\
			\emph{FreiHand}~\cite{zimmermann2019freihand}      & 21 &  &    &  \checkmark & & & 130K \\
			\emph{MHP}~\cite{gomez2019large}         & 21  &  &    & \checkmark & & & 80K \\
			\midrule
			\emph{WFLW}~\cite{wu2018look}      &  98 & \checkmark  &   &   & \checkmark & & 10K \\
			\emph{AFLW}~\cite{koestinger2011annotated}      & 19 & \checkmark  &   &   & \checkmark& & 25K \\
			\emph{COFW}~\cite{burgos2013robust}     & 29 & \checkmark  &   &   & \checkmark & &  1852 \\
			\emph{300W}~\cite{sagonas2016300}            & 68 & \checkmark &   &   & \checkmark & & 3837      \\ \midrule
			COCO-WholeBody~\cite{jin2020whole} & 133 & \checkmark  & \checkmark & \checkmark    & \checkmark & & 250K  \\ 
			\textbf{Halpe-FullBody} & \textbf{136} & \checkmark  & \checkmark & \checkmark    & \checkmark &\checkmark & 50K \\
			\bottomrule	
		\end{tabular}
	}
	\end{center}
	\caption{Overview of some popular public datasets for 2D keypoint estimation in RGB images. Kpt stands for keypoints, and \#Kpt means the annotated number. ``Wild'' denotes whether the dataset is collected in-the-wild. ``HOI'' denotes human-object-interaction body-part labels.}
	\label{tab:dataset}
	\vspace{-0.2in}
\end{table}

\subsection{Network}
For our two-step framework, various human detector and pose estimator can be adopted.

In the current implementation, we adopt off-the-shelf detectors include YOLOV3~\cite{redmon2018yolov3} and EfficientDet~\cite{tan2020efficientdet} trained on COCO~\cite{MSCOCO} dataset. We do not retrained these models as their released model already work well in our case.

For the pose estimator, we design a new backbone named FastPose, which yields both high accuracy and efficiency. The network structure is illustrated in Fig.~\ref{fig:fastpose}. We use ResNet~\cite{he2016deep} as our network backbone to extract features from the input cropped image. Three Dense Upsampling Convolution (DUC)~\cite{wang2018understanding} modules are adopted to upsample the extracted features, followed by a $1 \times 1$ convolution layer to generate heatmaps. The DUC module first applies 2D convolution to the feature map with dimension $h \times w \times c$ and then reshapes it to $2h \times 2w \times c' $ via a PixelShuffle~\cite{shi2016real} operation. 

To further boost the performance, we also incorporate deformable convolution operator into our ResNet backbone following~\cite{dai2017deformable} to improve the feature extraction. Such network is named as FastPose-DCN.

\subsection{System}
AlphaPose is developed based on both PyTorch~\cite{paszke2019pytorch} and MXNet~\cite{chen2015mxnet}. Benefiting from the flexibility of PyTorch, AlphaPose supports both Linux and Windows system. AlphaPose is highly optimized for the purpose of easy usage and further development, as we decompose the training and testing pipeline into different modules and one can easily replace or update different modules for custom purpose.  For the data loading module, we support image input by specifying image name, directory or a path list. Video file or stream input from camera are also supported. For the detection module, we adopt YOLOX~\cite{yolox2021} YOLOV3-SPP~\cite{redmon2018yolov3}, EfficientDet~\cite{tan2020efficientdet} and JDE~\cite{wang2019towards}. Detecting results from other detectors are also supported as a file input. Other trackers like~\cite{pang2020tubetk} can also be incorporated. For the data transform module, we implement vanilla box NMS and soft-NMS~\cite{bodla2017soft}. For the pose estimation module, we supports SimplePose~\cite{xiao2018simple}, HRNet~\cite{sun2019deep}, and our proposed FastPose with different variants like FastPose-DCN.  Our Re-ID based tracking algorithm is also available in this module. For the post processing module, we provide our parametric pose NMS and the OKS-based NMS~\cite{papandreou2017towards}. Another tracker PoseFlow~\cite{xiu2018pose} is available here and we support rendering for images and video. Our saving format is COCO format by default and can be compatible with OpenPose~\cite{cao2018openpose}. One can easily run AlphaPose with different setting by simply specifying the input arguments.

\begin{figure}[hbt]
\centering
\begin{tabular}{@{\hspace{0mm}}c@{\hspace{1mm}}c@{\hspace{1mm}}c@{\hspace{1mm}}c@{\hspace{1mm}}c}
\includegraphics[width=0.9\linewidth]{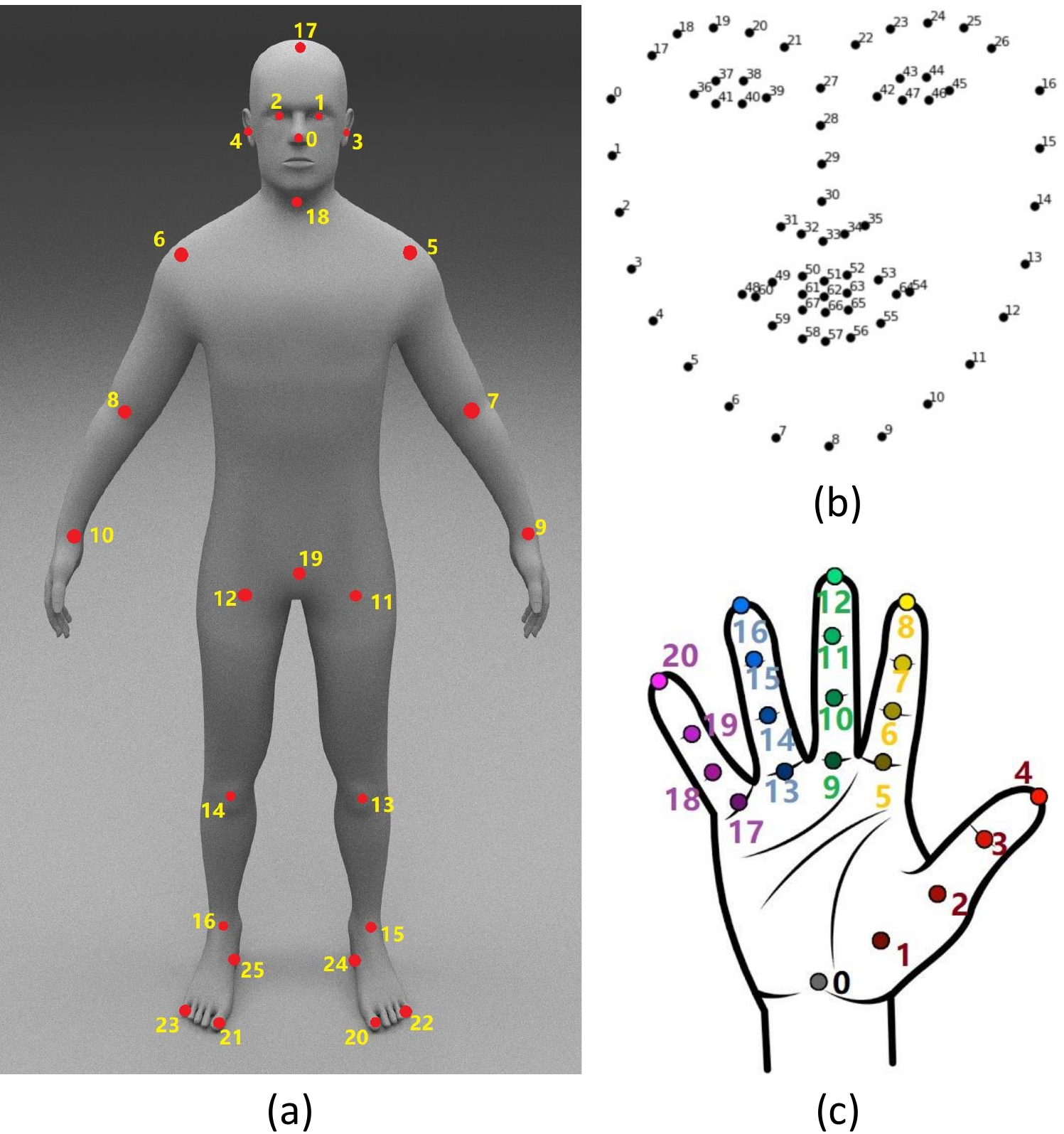}
\end{tabular}
\vspace{-0.1in}
\caption{Annotated keypoint format in Halpe-FullBody dataset for (a) body and foot, (b) face, (c) hand respectively. Zoom in for details of the face annotation,}
\label{fig:format}
\end{figure}

\begin{figure}[tb!]
\centering
\subfigure[]{
    \includegraphics[width=0.45\linewidth]{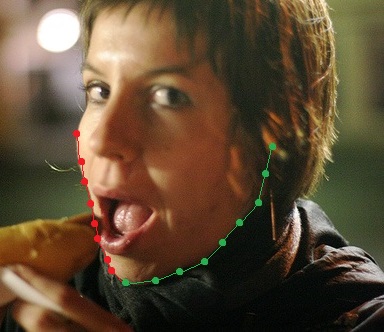}
}
\subfigure[]{
    \includegraphics[width=0.45\linewidth]{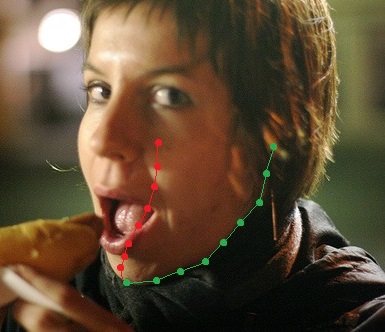}
}
\vspace{-0.2in}
\caption{Two different definition of face keypoints on the lower jaw. The green dots represent the same definition and the red dots indicate their differences. In (a) and (b), the left definition is commonly used in 2D annotated dataset like~\cite{wu2018look, sagonas2016300, jin2020whole}, while the right definition is used in 3D face alignment task like~\cite{zhu2016face}. }
\label{fig:face_define}
\end{figure}

\begin{table*}[t]
	\begin{center}
    \resizebox{\linewidth}{!}{
		\begin{tabular}{l|c|cccccc|cc|cc|cc|cc}
			\toprule
			Method & Input Size & \multicolumn{6}{c|}{full-body}  & \multicolumn{2}{c|}{foot}  & \multicolumn{2}{c|}{face}  & \multicolumn{2}{c|}{hand} & \multicolumn{2}{c}{body} \\
			\cmidrule{3-16}
			&  &  AP  &  AP$^{50}$  &  AP$^{75}$ &  AP$^{L}$ &  AP$^{M}$  & AR     & AP   & AR     &  AP  & AR     & AP    & AR   &  AP     & AR  \\
			\midrule
			OpenPose-default~\cite{cao2018openpose} & N/A & 0.276 &0.528 &0.258 &0.356 &0.310 & 0.370 & 0.438 & 0.652 & 0.482 & 0.495 & 0.140 & 0.209 & 0.514 & 0.575 \\
			OpenPose-maxacc~\cite{cao2018openpose} & N/A & 0.281 &0.531 &0.265 &0.363 &0.318 & 0.381 & 0.456 & 0.677 & 0.482 & 0.496 & 0.142 & 0.211 & 0.526 & 0.590 \\
			SN~\cite{hidalgo2019single} & N/A & 0.233 &0.606 &0.128 &0.211 &0.354 & 0.362 & 0.481 & 0.680 & 0.344 & 0.419 & 0.030 & 0.071 & 0.563 & 0.624 \\ 
			\midrule
			HRNet~\cite{sun2019deep}  & 256$\times$192 & 0.387 &0.782 &0.346 &0.393 &0.432 & 0.522 & 0.581 & 0.749 & 0.429 & 0.558 & 0.104 & 0.204 & 0.605 & 0.713 \\
			Simple~\cite{xiao2018simple} & 256$\times$192 & 0.409 &0.782 &0.391 &0.417 &0.435 &0.506 &0.706 &0.782 &0.444 &0.536 &0.141 &0.233 &0.648 &0.691  \\ 
			ZoomNet & 384$\times$288 & 0.427 & \textbf{0.803} & 0.412 &0.446 &0.433 &0.513 &0.702 &0.778 & 0.505 & 0.569 & 0.136 & 0.210 & 0.648 & 0.699  \\
			\midrule
			FastPose50-hm & 256$\times$192 & 0.417 &0.784 &0.406 &0.426 &0.439 & 0.516 & 0.730 & 0.803 & 0.432 & 0.536 & 0.163 & 0.258 & 0.658 & 0.701  \\
			FastPose50-si & 256$\times$192 & 0.441 &0.772 &0.444 &0.470 &0.446 & 0.532 & 0.706 & 0.781 & 0.491 & 0.580 & 0.207 & 0.294 & 0.650 & 0.699  \\
			FastPose152-si & 256$\times$192 & 0.451 &0.785 &0.457 &0.475 &0.460 & 0.537 & 0.724 & 0.791 & \textbf{0.508} & \textbf{0.590} & 0.199 & 0.294 & 0.651 & 0.699  \\
			FastPose50-dcn-si & 256$\times$192 & \textbf{0.462} &0.795 &\textbf{0.477} &\textbf{0.491} &\textbf{0.464} & \textbf{0.548} & \textbf{0.739} & \textbf{0.810} & 0.508 & 0.589 & \textbf{0.214} & \textbf{0.301} & \textbf{0.672} & \textbf{0.717}  \\
			FastPose50-dcn-si* & 256$\times$192 & \textbf{0.484} & \textbf{0.826} &\textbf{0.505} &\textbf{0.497} &\textbf{0.508} & \textbf{0.565} & \textbf{0.733} & \textbf{0.810} & \textbf{0.537} & \textbf{0.596} & \textbf{0.226} & \textbf{0.330} & \textbf{0.678} & \textbf{0.721}  \\
			\bottomrule
		\end{tabular}
		}
	\end{center}
	\caption{Whole-body pose estimation results on Halpe-FullBody dataset. For fair comparisons, results are obtained using single-scale testing. ``OpenPose-default'' and ``OpenPose-maxacc'' denotes its default and maximum accuracy configuration respectively. ``hm'' denotes the network uses heatmap based localization, ``si'' denotes the network uses our symmetric integral regression. ``*'' denotes model trained with multi-domain knowledge distillation and PGPG.\red{FastPose50 denotes our FastPose network with ResNet50 as backbone and so is FastPose152. ``dcn'' denotes that the deformable convolutional layer~\cite{dai2017deformable} is adopted in the ResNet backbone.}}
	\label{tab:compare_halpe}
\end{table*}

\begin{table*}[t]
	\begin{center}
		\begin{tabular}{c|l|c|c|cc|cc|cc|cc|cc}
			\toprule
			Type & Method & Input Size & GFLOPs & \multicolumn{2}{c|}{whole-body} & \multicolumn{2}{c|}{body}  & \multicolumn{2}{c|}{foot}  & \multicolumn{2}{c|}{face}  & \multicolumn{2}{c}{hand} \\
			\cmidrule{5-14}
			& &   &   &  AP     & AR     & AP   & AR     &  AP  & AR     & AP    & AR   &  AP     & AR  \\
			\midrule
			& OpenPose~\cite{cao2018openpose} & N/A & N/A & 0.338 & 0.449 &0.563 & 0.612 & 0.532 & 0.645 & 0.482 & 0.626 & 0.198 & 0.342  \\ 
			& SN~\cite{hidalgo2019single} & N/A & N/A & 0.161 & 0.209 & 0.280 & 0.336 & 0.121 & 0.277 & 0.382 & 0.440 & 0.138 & 0.336  \\ 
			Bottom-& PAF~\cite{cao2017realtime} & N/A & N/A & 0.141 & 0.185 & 0.266 & 0.328 & 0.100 & 0.257 & 0.309 & 0.362 & 0.133 & 0.321  \\ 
			Up& PAF-body~\cite{cao2017realtime} & N/A & N/A & - & - & 0.409 & 0.470 & - & - & - & - & - & -  \\ 
			& AE~\cite{newell2017associative} & N/A & N/A & 0.274 & 0.350 & 0.405 & 0.464 & 0.077 & 0.160 & 0.477 & 0.580 & 0.341 & 0.435  \\
			& AE-body~\cite{newell2017associative} & N/A & N/A & - & - & 0.582 & 0.634 & - & - & - & - & - & -  \\ 
			\midrule
			& HRNet~\cite{sun2019deep}  & 384$\times$288 & 16.0 & 0.432 & 0.520 & 0.659 & 0.709 & 0.314 & 0.424 & 0.523 & 0.582 & 0.300 & 0.363  \\
			& HRNet-body~\cite{sun2019deep} & 384$\times$288 & 16.0 & - & - & \textbf{0.758} & \textbf{0.809} & - & - & - & - & - & -   \\ 
			Top- & ZoomNet & 384$\times$288 & 20.0 & 0.541 & 0.658 & 0.743 & 0.802 & \textbf{0.798} & \textbf{0.869} & 0.623 & 0.701 & 0.401 & 0.498   \\
			\cmidrule{2-14}
			Down& FastPose50-si & 256$\times$192 & 5.9 & 0.554 & 0.625 & 0.673 & 0.717 & 0.636 & 0.718 & 0.757 & 0.818 & 0.425 & 0.515  \\
			& FastPose152-si & 256$\times$192 & 13.2 & 0.569 & 0.641 & 0.684 & 0.730 & 0.672 & 0.750 & \textbf{0.765} & \textbf{0.824} & 0.443 & 0.532  \\ 
			& FastPose50-dcn-si & 256$\times$192 & 6.1 & \textbf{0.577} & \textbf{0.650}  & 0.693 & 0.740 & 0.690 & 0.765 & 0.759 & 0.820 & \textbf{0.453} & \textbf{0.538} \\
			\bottomrule
		\end{tabular}
	\end{center}
	\caption{Whole-body pose estimation results on COCO-WholeBody dataset. For fair comparisons, results are obtained using single-scale testing. We only report the input size and GFLOPS of the pose model in top-down based approaches and ignore the detection model.``hm'' denotes the network uses heatmap based localization, ``si'' denotes the network uses our symmetric integral regression. \red{FastPose50 denotes our FastPose network with ResNet50 as backbone and so is FastPose152. ``dcn'' denotes that the deformable convolutional layer~\cite{dai2017deformable} is adopted in the ResNet backbone.}}
	\label{tab:compare_cocowhole}
\end{table*}
	
\begin{table*}[t]
	\begin{center}
    \resizebox{\linewidth}{!}{
		\begin{tabular}{c|l|c|c|c|c|c|ccccc}
		    \toprule
			\multicolumn{2}{c|}{Methods} & Backbone & Detector & Input Size & GFLOPs & AP & AP$^{50}$  &  AP$^{75}$ &  AP$^{M}$ &  AP$^{L}$  & AR  \\
			\midrule
			\multirow{7}{*}{Detection} & G-MRI~\cite{papandreou2017towards} & ResNet-101 & Faster-RCNN & $353 \times 257$ & 57.0 & 0.649 & 0.855 & 0.713 & 0.623 & 0.700 & 0.697 \\ 
			&  RMPE~\cite{fang2017rmpe} & PyraNet & Faster-RCNN & $320 \times 256$ & 26.7 & 0.723 & 0.892 & 0.791 & 0.680 & 0.786 & - \\ 
			&  CPN~\cite{cao2017realtime} & ResNet-Inception & FPN & $384 \times 288$ & - & 0.721 & 0.914 & 0.800 & 0.687 & 0.772 & 0.785 \\ 
			&  PAF-body~\cite{cao2017realtime} & - & - & - & - & 0.618 & 0.849 & 0.675 & 0.571 & 0.682 & 0.665 \\ 
			&  AE~\cite{newell2017associative} & - & - & - & - & 0.655 & 0.868 & 0.723 & 0.606 & 0.726 & 0.702 \\
			& SimplePose~\cite{xiao2018simple} & ResNet-50 & Faster-RCNN & $256 \times 192$ & 8.9 & 0.702 & 0.909 & 0.783 & 0.671 & 0.759 & 0.758 \\
			&  HRNet~\cite{sun2019deep}  & HRNet-32 & Faster-RCNN & $384 \times 288$ & 16.0 & 0.749 & 0.925 & 0.828 & 0.713 & 0.809 & 0.801 \\
            &  \red{HRNet}~\cite{sun2019deep}  & HRNet-48 & Faster-RCNN & $384 \times 288$ & 32.9 &  0.755  & 0.925 &  0.833 &  0.719 &  0.815 &  0.805 \\
			\cmidrule{2-12}
			& FastPose-hm & ResNet-50 & YOLO-v3 & $256 \times 192$ & {5.9} & 0.718 & 0.919 & {0.803} & {0.728} & {0.742} & {0.773} \\
			& FastPose-dcn-hm & ResNet-50 & YOLO-v3 & $256 \times 192$ & {6.1} & 0.726 & 0.922 & {0.812} & {0.737} & {0.749} & {0.781} \\
			& FastPose-dcn-hm & ResNet-101 & YOLO-v3 & $256 \times 192$ & {9.8} & 0.727 & 0.922 & {0.813} & {0.736} & {0.751} & {0.781} \\
			\midrule
			\multirow{5}{*}{Regression}
			&  Integral~\cite{sun2018integral} & ResNet-101 & Faster-RCNN & $256 \times 256$ & 17.8 & 0.678 & 0.882 & 0.748 & 0.639 & 0.740 & - \\
			& CenterNet~\cite{xingyi2019object} & Hourglass-2 stacked & - & - & - & 0.630 & 0.868 & 0.696 & 0.589 & 0.704 & - \\
			& SPM~\cite{nie2019single} & Hourglass-8 stacked & - & - & - & 0.669 & 0.885 & 0.729 & 0.626 & 0.731 & - \\
			& Point-set Anchor~\cite{wei2020point} & HRNet-W48 & - & - & - & 0.687 & 0.899 & 0.763 & 0.648 & 0.753 & - \\
			\cmidrule{2-12}
			&  FastPose-si & ResNet-50 & YOLO-v3 & $256 \times 256$ & {7.9} & 0.649 & 0.865 & {0.728} & {0.669} & {0.663} & {0.716} \\
			&  FastPose-si & ResNet-101 & YOLO-v3 & $256 \times 256$ & {12.8} & 0.679 & 0.876 & {0.751} & {0.675} & {0.714} & {0.723} \\
			&  FastPose-dcn-si & ResNet-101 & YOLO-v3 & $256 \times 256$ & {13.1} & {0.690} & 0.901 & {0.773} & {0.729} & {0.690} & {0.775} \\
			\bottomrule
		\end{tabular}
	}
	\end{center}
	\caption{Body pose estimation results on COCO test-dev set. For fair comparisons, results are obtained using single-scale testing. ``hm'' denotes the network uses heatmap based localization, ``si'' denotes the network uses our symmetric integral regression.}
	\label{tab:compare_coco}
\end{table*}

\section{Datasets and Evaluations} 
\subsection{Datasets}
\label{sec:datasets}
\noindent{\bf Halpe-FullBody} To facilitate the development of whole body human pose estimation, we annotate a full body keypoints dataset named \textbf{Halpe-FullBody}\footnote{Available at \url{https://github.com/Fang-Haoshu/Halpe-FullBody}}. For each person, we annotate 136 keypoints, including 20 for body, 6 for feet, 42 for hands and 68 for face. The keypoint format is illustrated in Fig.~\ref{fig:format}. Note that since there are two popular definition for the face keypoints (see Fig.~\ref{fig:face_define}), we only annotate the visible lower jaw of the face (green dots in Fig.~\ref{fig:face_define}) so as to be compatible with these two definition. For the images, our training set uses the training images of the HICO-DET~\cite{chao2018learning} dataset and our testing set uses the COCO-val set. In total, our dataset contains 50K instances for training and 5K images for testing. Tab.~\ref{tab:dataset} compares our dataset with previous popular datasets on human pose estimation.

\noindent{\bf COCO-WholeBody}
As a concurrent work, Jin \textit{et. al.} annotates 133 whole body keypoints based on the COCO dataset. They share a similar keypoints definition with us, except that the head, neck and hip points are missing in their annotation. The total training set contains 118K images with 250K instances, and the test set contains 5K images. We also evaluate our algorithm on this dataset.

\noindent{\bf COCO}
COCO dataset is a standard benchmark for human keypoints prediction. It contains 17 keypoints of human body without face, hand and foot annotations. In total, there are 118K images for training, 5K for validation and 41K for testing. We \red{train our algorithm on the COCO 2017 train set and} compare our FastPose network and symmetrical integral loss with previous state-of-the-arts models \red{on the COCO 2017 test-dev set.}

\noindent{\bf PoseTrack}
PoseTrack is a large scale dataset for multi-person pose estimation and tracking. It is built on the raw videos provided by MPII Human dataset\cite{andriluka14cvpr}. There are more than 1356 video sequences of PoseTrack and they are split into train, val, test. Each annotated person has 17 keypoints similar with COCO, but there are two different keypoints compared with COCO, which are ‘top head’ and ‘bottom head’. Other annotations share the same format with COCO. \red{We train our method on PoseTrack-2018 set and compare it with previous methods on both PoseTrack-2017-val and PoseTrack-2018-val sets.}

\noindent{\bf 300Wface, FreiHand and InterHand} are used as supplemental datasets to improve the generalization ability of our model. 300Wface~\cite{sagonas2016300} contains 300 indoor and 300 outdoor in-the-wild images. For each face, 68 keypoints are annotated. FreiHand~\cite{zimmermann2019freihand} contains 33K unique hand samples for training, each contains 21 keypoints. InterHand~\cite{moon2020interhand2} contains 2.6M images with interacting hands, where each hand also has 21 keypoints.

\subsection{Evaluation Metrics and Tools}
\noindent{\bf Halpe-FullBody}
We extend the evaluation metric of the COCO keypoints to full body scenario. COCO defines a Object Keypoint Similarity controlled by a per-keypont constant $k$. For our newly added keypoints, we set the $k$ for the feet, face and hand as 0.015. Same as COCO, we report $AP^{0.5:0.95:0.05}$ as the main result and the detailed results for body, foot, face and hand are also reported.

\noindent{\bf COCO-WholeBody} COCO-WholeBody adopts the same metric as ours except that the constant $k$ is different from us for some keypoints.

\noindent{\bf COCO} We adopt the standard AP metric of COCO dataset for fair comparison with previous works.

\noindent{\bf PoseTrack}
In fact, multi-person pose tracking can be regarded as the combination of multi-person pose estimation and multi-object tracking. Thus, the evaluation metric should follow these two tasks. Mean Average Precision (mAP)\cite{pishchulin16cvpr} is used to measure frame-wise human pose accuracy. To evaluate the tracking performance, the MOT\cite{milan2016mot16} metric is applied to each of the body joints independently. Then the final 
tracking performance is obtained by averaging each joint mot metric. The PCKh\cite{andriluka14cvpr} (head-normalized probability of correct keypoint) is one of the most commonly used metric to evaluate whether a body joint is predicted correctly. Here it can determine which predicted joint is matched with groundtruth joint.

To evaluate the tracking result on posetrack validation dataset, we use the official tool named poseval\footnote{https://github.com/leonid-pishchulin/poseval} and report Multiple Object Tracker Accuracy (MOTA), Multiple Object Tracker Precision (MOTP), Precision and Recall.

\subsection{Implementation Details}
We conduct our experiments with PyTorch~\cite{paszke2019pytorch}.  We train the network with batch 32 for 270 epochs. The initial learning rate is 0.01 and we decay it on epoch 100 and 170 by 0.1. The pose guided proposal generator is applied after epoch 200. After the entire network is trained, we freeze the backbone and only finetune the \redred{re-ID} branch on posetrack dataset for 10 epochs. Learning rate in the finetuning phase is 1e-4. We adopt Adam~\cite{kingma2014adam} optimizer during training. All experiments are conducted on 8 Nvidia 2080Ti GPUs.

\subsection{Evaluation for Full Body Pose Estimation}
We first evaluate the performance of our model on Halpe-FullBody and COCO-WholeBody dataset. Since Halpe-FullBody is a new dataset, we retrain several state-of-the-art models and compare the results with us. Tab.~\ref{tab:compare_halpe} gives the final results. YOLOV3 is adopted as human detector for all the top-down based models. We can see that top-down methods can achieve higher accuracy compared to the bottom-up methods. However, due to the quantization error introduced by heatmap, conventional SPPEs decrease a lot on the fine-level body parts like face and hand. Equipped by our novel symmetrical integral loss function, our FastPose models achieve the best accuracy. \red{Notably, we can see that FastPose50-si yields 2.4 mAP (5.7\% relatively) higher than its heatmap-based counterpart. The improvements mainly comes from the face and hands. It demonstrates that the quantization error of heatmap affects the fine-level localization of face and hand keypoints, and our symmetric integral regression works well on such cases.}

\begin{figure*}[hbt]
\centering
\begin{tabular}{@{\hspace{0mm}}c@{\hspace{1mm}}c@{\hspace{1mm}}c@{\hspace{1mm}}c@{\hspace{1mm}}c}
\includegraphics[width=\linewidth]{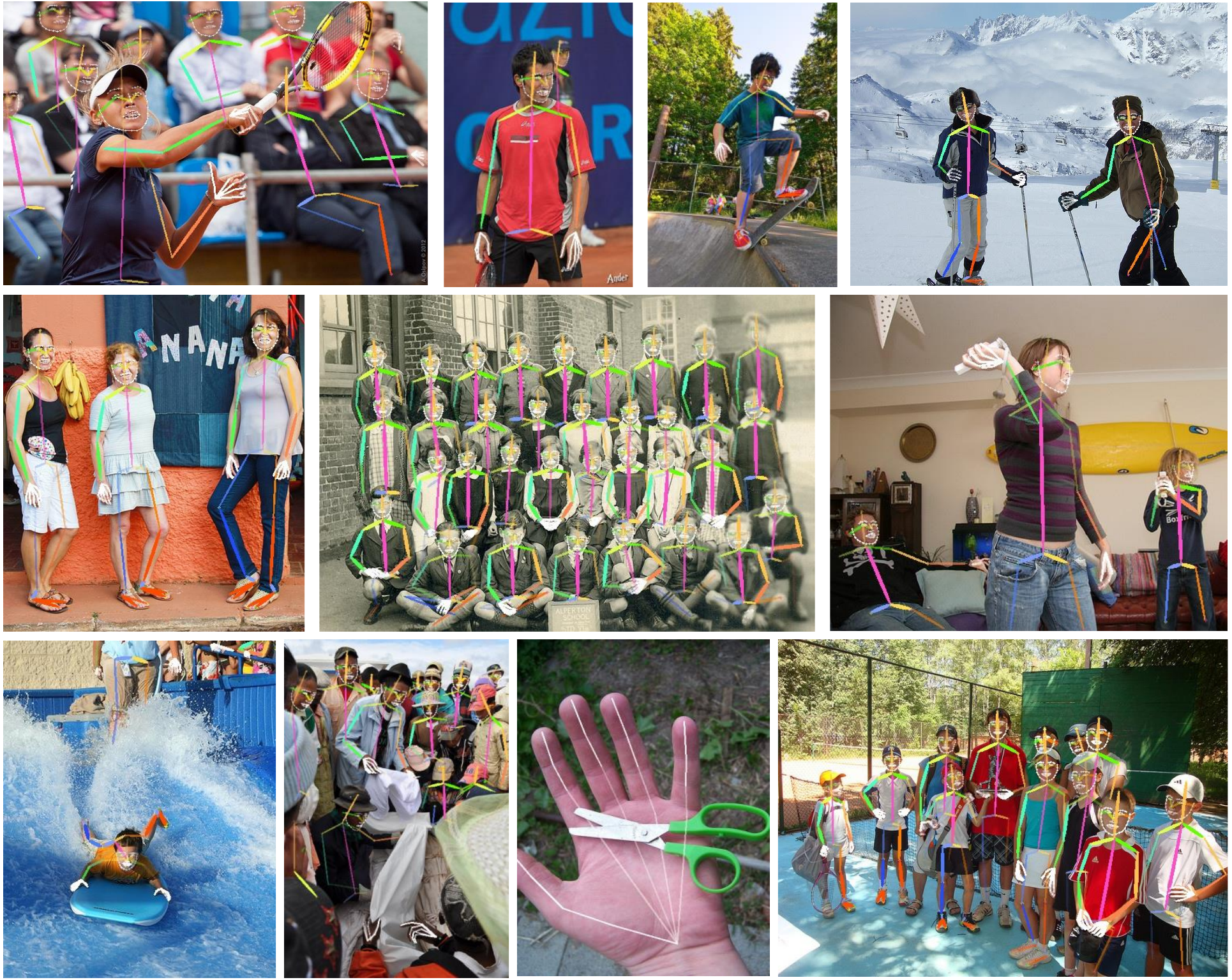}
\end{tabular}
\caption{Qualitative results of AlphaPose on the full-body pose estimation task. Zoom in for more details and best viewed in color. }
\label{fig:res}
\end{figure*}

\begin{figure*}[hbt]
\centering
\begin{tabular}{@{\hspace{0mm}}c@{\hspace{1mm}}c@{\hspace{1mm}}c@{\hspace{1mm}}c@{\hspace{1mm}}c}
\includegraphics[width=\linewidth]{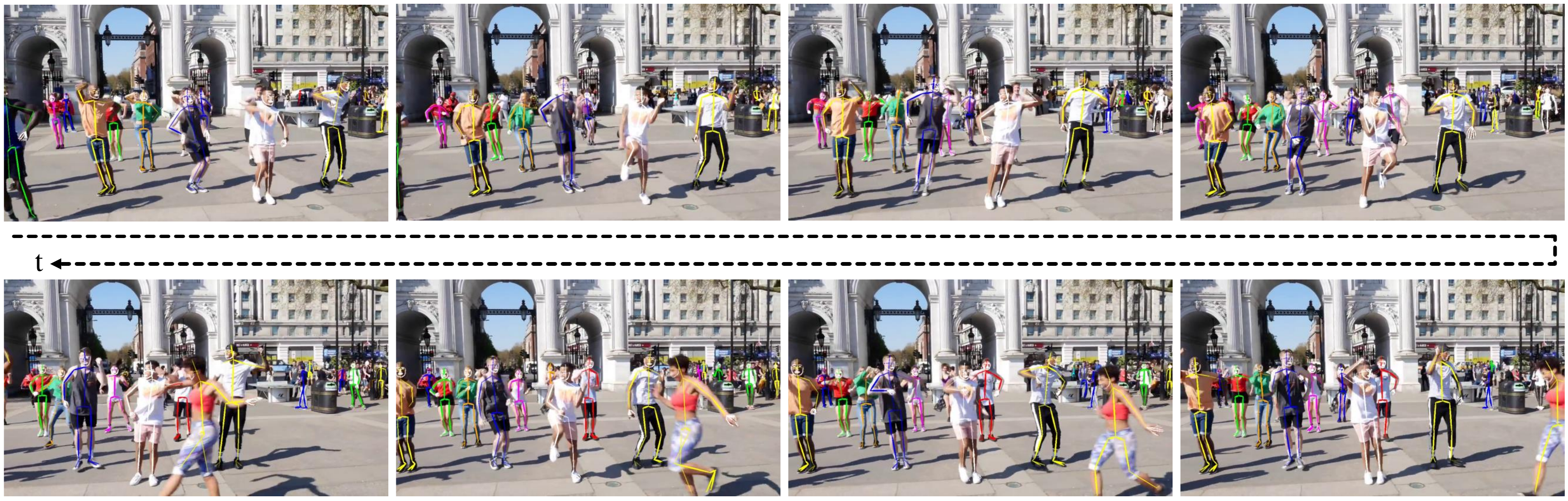}
\end{tabular}
\vspace{-0.18in}
\caption{Qualitative results of AlphaPose on the full-body pose tracking task. Zoom in for more details and best viewed in color. The colors of persons denote their tracking ID. \red{The image order is denoted by the time arrow. See text for more analysis.}}
\label{fig:res_tracking}
\end{figure*}

On the COCO-WholeBody dataset, our FastPose embedded with symmetrical integral loss function also outperforms previous state-of-the-art methods by a large margin, especially on the face and hands. Notably, our FastPose achieves the highest accuracy given a smaller input size. The model complexity is also much lower than previous methods. It demonstrates the superiority of our network structure and the novel loss. 

Some qualitative results of full body pose estimation is shown in Fig.~\ref{fig:res}.

\subsection{Evaluation for Conventional Body Pose Estimation}
\red{We also conduct experiments on the conventional body-only pose estimation task to demonstrate the effectiveness of our method, although it is not our main focus.}. We train our models on the COCO dataset and evaluate it on COCO test-dev set. The results are reported in Tab.~\ref{tab:compare_coco}.  For the heatmap based methods, we can see that our FastPose backbone can achieve on par performance with the state-of-the-art method, given a \emph{smaller input size} and \emph{weaker human detector}. It demonstrates the superiority of our FastPose network. \red{Note that since our goal is to present a new baseline model like SimplePose~\cite{xiao2018simple}, we conduct these experiments to prove the accuracy and efficiency of our model. Further pursuing higher accuracy with speed and resources trade-off is not our goal in this paper and we leave them for future research.} 

For the regression based methods, our method achieves the state-of-the-art performance with the lowest GFLOPS. \red{Compared to~\cite{sun2018integral}, our network serves as a new baseline for future research.}

\begin{table}
    \begin{center}
    \begin{tabular}{l|c|c}
    \toprule
    Module & Halpe Fullbody (mAP) &  COCO (mAP)  \\
    \midrule
    \redred{two-step hm-norm} &  44.1 & 69.5 \\
    \redred{one-step hm-norm} & 38.1 & 67.1\\
    \cmidrule{1-3}
    w. SIKR &  44.1 & 69.5 \\
    w.o SIKR & 42.3 & 64.6\\
    \cmidrule{1-3}
    w. P-NMS &44.1  &69.5 \\
    w.o P-NMS &43.7 & 68.2 \\
    \cmidrule{1-3}
    w. PGPG & 48.4* &  N/A \\
    w.o PGPG & 47.1* &  N/A \\
    \bottomrule
    \end{tabular}
    \end{center}
    \caption{Ablation studies on Halpe Fullbody dataset and COCO dataset. ``hm-norm'' denotes heatmap normalization. ``*'' denotes results trained with additional data from Multi-Domain Knowledge Distillation.}
    \label{tab:pose_ablation}
    \vspace{-0.2in}
\end{table}

\subsection{Ablation Studies for Pose Estimation}
\label{sec:ablation}
To evaluate the effectiveness of our proposed module for pose estimation, we also conducted ablation experiments on COCO and Halpe-Fullbody dataset. We adopt FastPose50 as base network and report the numbers on COCO validation and Halpe-Fullbody test set respectively. The results are summarized in Tab.~\ref{tab:pose_ablation}.

\noindent{\bf Heatmap Normalization} \redred{We elucidated the essence of our two-step heatmap normalization for applying the integral-based method in the multi-person scenario in Sec.\ref{sec:SIKR}. Here we conduct an ablation experiment to show the performance gap of different heatmap normalization methods. We can see that when comparing the conventional one-step heatmap normalization (soft-max) to our two-step heatmap normalization, the performance in multi-person pose estimation decrease for 6 mAP and 2.4 mAP on Halpe-fullbody and COCO datasets, respectively. It demonstrates that the two-step normalization can alleviate the size-dependent effect and improve performance.}

\noindent{\bf SIKR Module} We compare our symmetric integral function with the original integral regression~\cite{sun2018integral}. For both full body pose estimation scenario and conventional body pose estimation, we can see that our symmetric integral function greatly outperforms the original integral regression.

\noindent{\bf Pose NMS Module} Without Pose-NMS, multiple human poses will be predicted for a single person. The redundant poses will decrease the model performance. From Tab.~\ref{tab:pose_ablation}, we can see that our model decreases for 0.4 mAP and 1.3 mAP for Halpe Fullbody and COCO dataset respectively.

\noindent{\bf PGPG Module} Proper data augmentation is needed during training to ensure the generalization ability at testing phase. For Halpe Fullbody dataset, we compare the results of FastPose50-dcn trained with and without PGPG module. Tab.~\ref{tab:pose_ablation} shows that without our part guided proposal generation, the performance would decrease due to the domain variance in training.

\subsection{Evaluation for Pose Tracking}
To verify that our system is sufficient for multi-person pose tracking task, we apply it to the posetrack validation dataset. Tab. \ref{tab:qua} shows the comparison with other state-of-the-art methods. The backbone we adopted is the FastPose152 and the detector is YoloX. We can see that our model outperforms most methods in both mAP metric and MOTA metric, and our speed is quite fast.  This near real-time processing speed can be applied to various scenarios in our real life. It is worth noting that there are some other methods\cite{girdhar2018detect,wang2020combining} that have achieved good results on the posetrack dataset, but they mainly consider the overall timing information of the video, which means that they are not strictly an online algorithms. Therefore our method is not directly compared with theirs. \red{\cite{bao2020pose, yang2021learning, snower202015} achieve higher accuracy compared with our results. But they use very high resolutions for input and output, which consumes a lot of memory and is computationally expensive. Our method achieves satisfactory accuracy while running efficiently.}
\begin{table}
    \begin{center}
    \begin{tabular}{l|c|c|c|c|c|c}
    \toprule
    Data & Method & mAP & MOTA & fps & \red{Res} & \red{Src}\\
    \midrule
    &Det\&Track\cite{girdhar2018detect} & 60.6  & 55.2 & 1.2 &-&\checkmark\\
    &PoseFlow\cite{xiu2018pose} & 66.5 & 58.3 & 10* &-&\checkmark\\
    &JointFlow\cite{doering2018joint} & 69.3  & 59.8 & 0.2 & N/A&$\times$\\
    &Fast\cite{zhang2019fastpose} & 70.3 & 63.2&12.2 &N/A&$\times$\\
    &TML++\cite{hwang2019pose} & 71.5 & 61.3 & - &-&$\times$\\
    &STAF\cite{raaj2019efficient} & 72.6 & 62.7 & 3.0 &N/A&\checkmark\\
    2017&FlowTrack\cite{xiao2018simple} & 76.7 & 65.4 & 3.0 &384$\times$288 & \checkmark\\
    &\red{PGPT\cite{bao2020pose}} & 77.2 & 68.4 & 1.2 & 384$\times$288 &\checkmark\\
    &\red{Yang~\textit{et.al.}\cite{yang2021learning}} & 81.1 & 73.4 & - & 384$\times$288 &$\times$\\
    &\textbf{Ours-UNI} & 76.1 & 65.5 & 11.3 & 256$\times$192 &\checkmark\\
    &\textbf{Ours-SEP} & \textbf{76.9} & \textbf{65.7} & 8.9 & 256$\times$192 &\checkmark\\
    \midrule
    &\red{MDPN\cite{guo2018multi}} & 71.7 & 50.6 & - &384$\times$288&$\times$\\
    &STAF\cite{raaj2019efficient} & 70.4  & 60.9 & 3.0 &N/A&\checkmark\\
    &OpenSVAI\cite{ning2018top} & 69.7 & 62.4 & - &-& $\times$\\
    &LightTrack\cite{ning2019lighttrack} & 71.2 & 64.6 & 0.7 &384$\times$288& \checkmark\\
    2018&\red{KeyTrack\cite{snower202015}} & 74.3 & 66.6 & 1.0 & 384$\times$288 & $\times$\\
    &\red{PGPT\cite{bao2020pose}} & 76.8 & 67.1 & 1.2 & 384$\times$288 & \checkmark\\
    &\red{Yang~\textit{et.al.}\cite{yang2021learning}} & 77.9 & 69.2 & - & 384$\times$288 & $\times$\\
    &\textbf{Ours-UNI} & 74.0 & 64.4 & \textbf{10.9} & 256$\times$192 & \checkmark\\
    &\textbf{Ours-SEP} & \textbf{74.7} & \textbf{64.7} & 8.7 & 256$\times$192 & \checkmark\\
    \bottomrule
    \end{tabular}
    \end{center}
    \caption{Evaluation Result On Posetrack Validation dataset. \red{``Res'' denotes input resolution of pose network and ``Src'' denotes whether source code is available.}  ``Ours-UNI'' denotes results trained with a shared backbone of pose and \redred{re-ID} branch and  ``Ours-SEP'' denotes results trained with separated backbones. The ``*'' in fps means not including detection time. The mAP value is obtained after tracking post-precessing.}
    \label{tab:qua}
    \vspace{-0.2in}
\end{table}

\subsection{Ablation Studies for Pose Tracking}
\label{sec:posetrack_ablation}
In order to verify the effectiveness of each part of the tracking algorithm, we have designed several sets of ablation experiments.

\begin{figure}[hbt]
\centering
\begin{tabular}{@{\hspace{0mm}}c@{\hspace{1mm}}c@{\hspace{1mm}}c@{\hspace{1mm}}c@{\hspace{1mm}}c}
\includegraphics[width=0.9\linewidth]{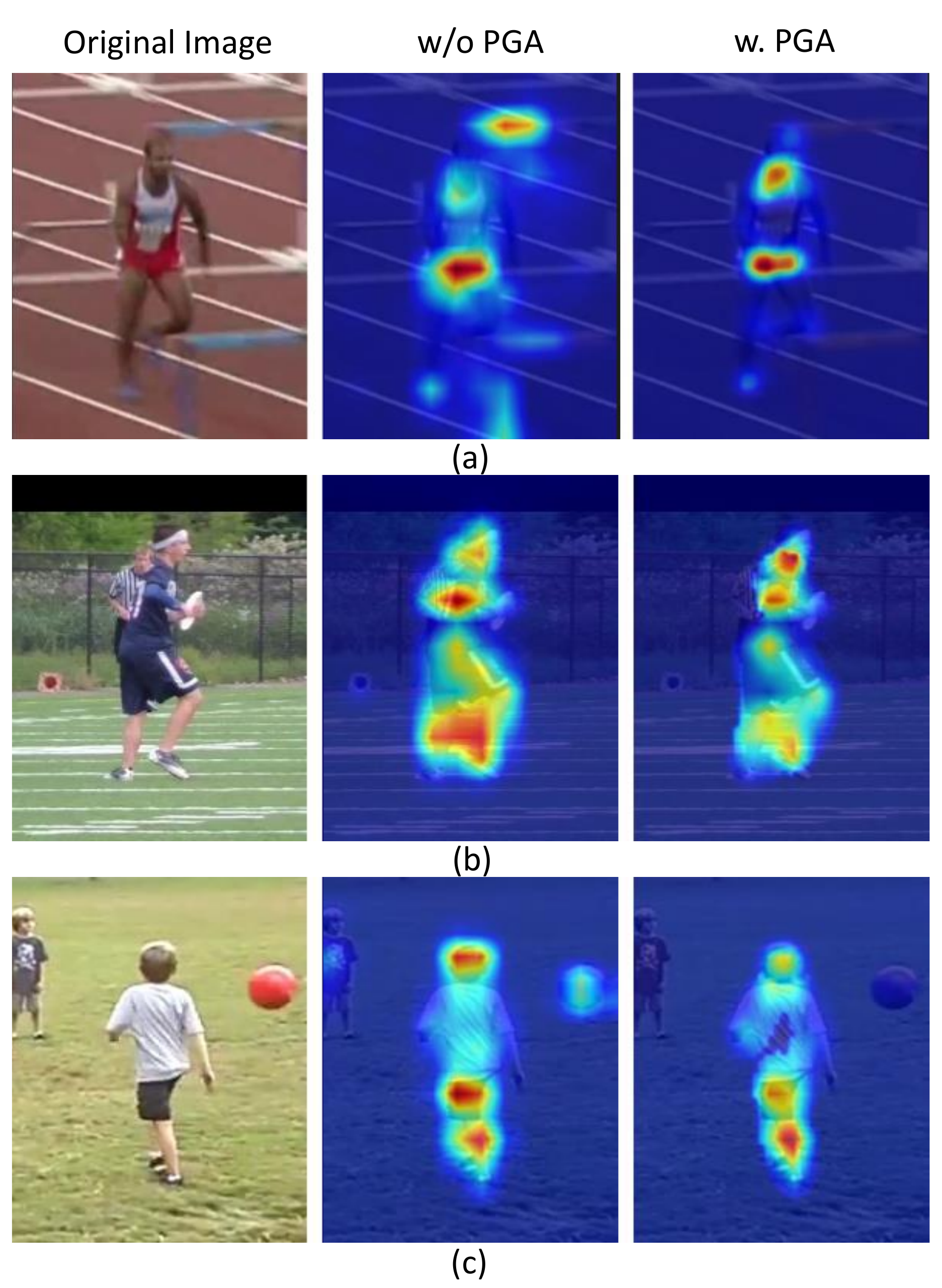}
\end{tabular}
\caption{Visualization of the role of PGA module. When there is no PGA module, some background area will also have a high response. The results of adding PGA module show that the feature response is more concentrated on the target person. \red{Notably, from figure (b) we can see that when two people are close, the feature response focus on the target person with the aid of PGA (zoom in for more details).}}
\label{fig:pga}
\end{figure}

\noindent{\bf PGA Module} The function of PGA module is to assist in extracting more effective \redred{re-ID} features with the help of the keypoint information. As a comparison, we remove the PGA module in our framework, which means the human pose and \redred{re-ID} feature are fed into MSIM directly. Tests on PoseTrack dataset show that tracking performance will decrease after removing PGA module which reported in Table.\ref{tab:PGA-MSIM}. At the same time, we visualized the extracted \redred{re-ID} features with or without PGA module shown as Fig.\ref{fig:pga}. Since the detection result is usually a larger box than the original size of human, the background in the box has a large proportion. However, the background information makes the human identity embedding carry useless features. This intuitively explains that the advantage of PGA is that it can better focus attention on the target person's area.
\begin{table}
    \begin{center}
    \resizebox{\linewidth}{!}{
    \begin{tabular}{c|c|c|c|c|c|c|c}
    \toprule
    exp&Setting & head & shou & elb&hip&knee& Total \\
    \midrule
    \multirow{6}*{PGA}& \multicolumn{7}{c}{mAP} \\
    \cmidrule{2-8}
    ~&w/ & 77.7  & 75.4 & 75.3&69.0&68.1&74.7\\
    \cmidrule{2-8}
    ~&w/o & 78.0  & 75.6 & 75.5&69.3&68.6&74.9\\
    \cmidrule{2-8}
    ~& \multicolumn{7}{c}{MOTA} \\
    \cmidrule{2-8}
    ~&w/  & 74.0  & 72.6 & 64.6 &66.8&58.9&64.7\\
    \cmidrule{2-8}
    ~&w/o& 73.4 & 71.8 & 64.2 &66.1&58.4&63.5\\
    \midrule
    
    \multirow{8}*{MSIM}& \multicolumn{7}{c}{mAP} \\
    \cmidrule{2-8}
    ~&No-GT & 77.7  & 75.4 & 75.3&69.0&68.1&74.7\\
    \cmidrule{2-8}
    ~&GT-Box& 81.3 & 81.0 & 81.5 &80.8&81.5&81.3\\
    \cmidrule{2-8}
    ~&GT-Pose& 100.0&100.0&100.0&100.0&100.0&100.0\\
    \cmidrule{2-8}
    ~& \multicolumn{7}{c}{MOTA} \\
    \cmidrule{2-8}
    ~&No-GT & 74.0  & 72.6 & 64.6 &66.8&58.9&64.7\\
    \cmidrule{2-8}
    ~&GT-Box& 75.8 & 75.5 & 75.8 &75.4&76.7&75.9\\
    \cmidrule{2-8}
    ~&GT-Pose& 93.8&93.6&93.7&93.8&94.4&93.9\\
    \bottomrule
    \end{tabular}
    }
    \end{center}
    \caption{The ablation study results of proposed pose tracking method. }
    \label{tab:PGA-MSIM}
    \vspace{-0.23in}
\end{table}

\noindent{\bf MSIM} To further verify the performance of our model, we add different level information into the Network. Specifically, we set up several sets of experiments, respectively using GT box, GT pose. These results are reported in Table.\ref{tab:PGA-MSIM}. The results show that if we replace the human detector and pose estimator with more accurate network, our tracking performance will be further improved.

\section{Full Body Pose Tracking}
In the above sections, we demonstrate the effectiveness of our methods on both full body pose estimation and pose tracking. Since our tracking algorithm is general, it is also applicable to the whole-body scenario. We adopt a weakly supervised strategy by training on both PoseTrack dataset and Halpe-FullBody dataset. 

Some qualitative results of full body pose tracking is shown in Fig.~\ref{fig:res_tracking}. \red{We can see that both full body pose estimation and pose tracking yield high accuracy given the heavily crowded scene. And our method is insensitive to the size variance of humans. Specifically, when a person is occluded by others and re-appear, our method still gives the correct identity (\emph{e.g.}, the person with black shorts on the right).}

\begin{figure}[bt]
\centering
\begin{tabular}{@{\hspace{0mm}}c@{\hspace{1mm}}c@{\hspace{1mm}}c@{\hspace{1mm}}c@{\hspace{1mm}}c}
\includegraphics[width=0.7\linewidth]{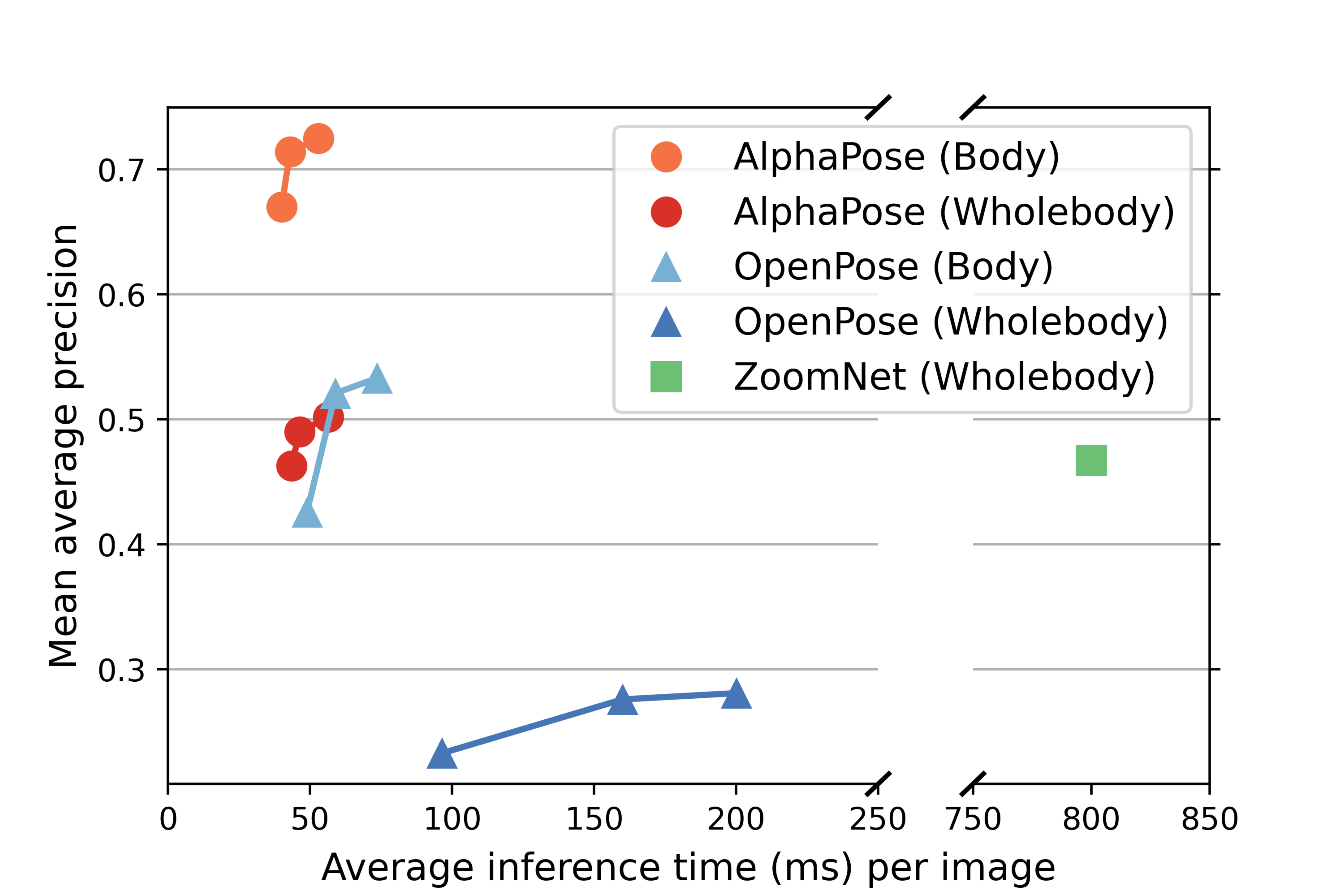}\\(a)\\
\includegraphics[width=0.7\linewidth]{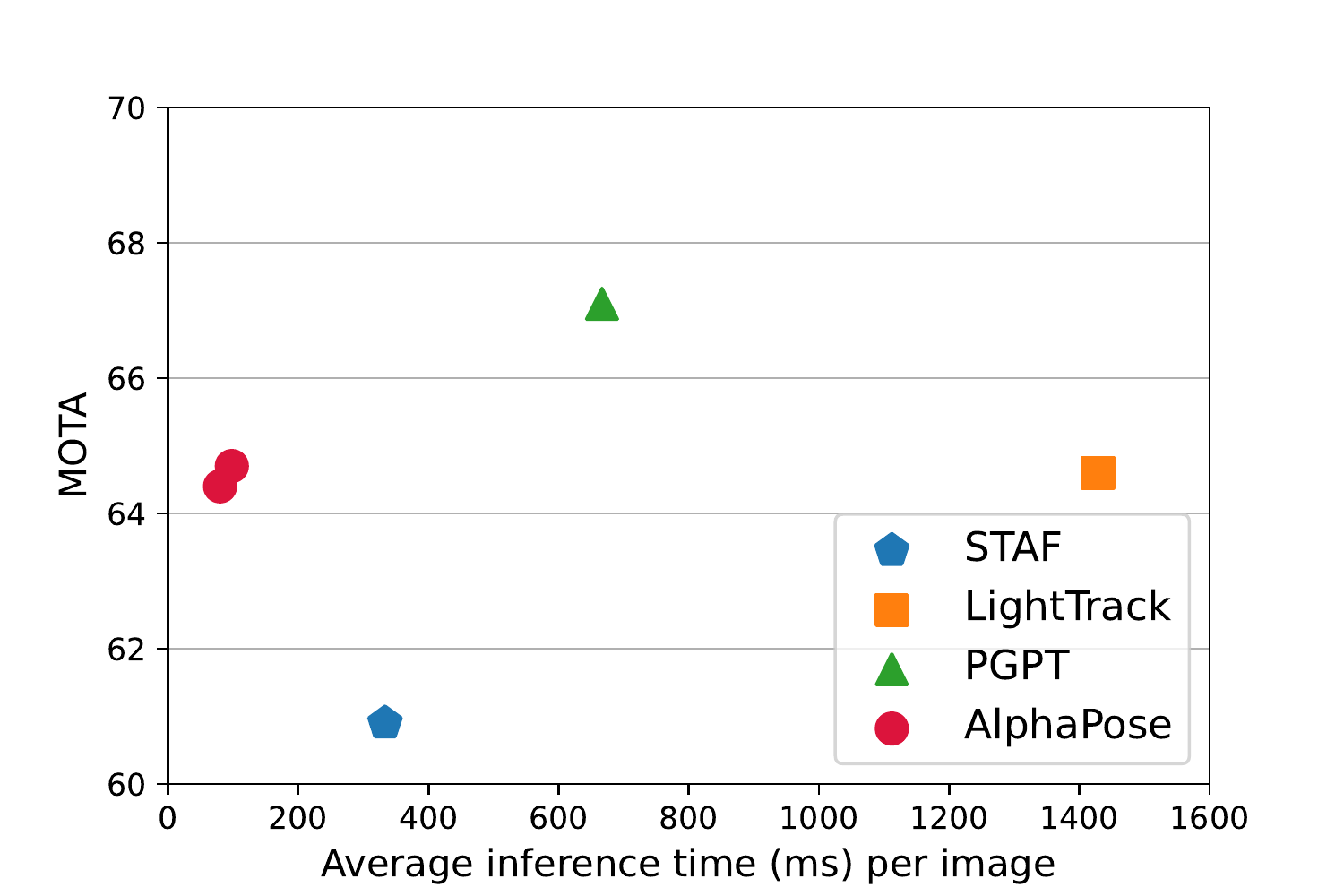}\\(b)
\end{tabular}
\caption{Speed/Accuracy comparison of different pose estimation and tracking libraries. \red{(a) Pose estimation results obtained on COCO-WholeBody validation set and COCO validation set. (b) Pose tracking results obtained on PoseTrack18-val set.}}
\label{fig:speedacc}
\end{figure}

\section{Library Analysis}
In this section, we compare our AlphaPose library with other popular open source library in both pose estimation and pose tracking. \red{The results are obtained on a single Nvidia 20080Ti GPU.} Fig.~\ref{fig:speedacc} shows the speed-accuracy curve of different libraries. From Fig.~\ref{fig:speedacc}(a) we can see that our method has the highest accuracy and yields the highest efficiency on whole-body and body-only pose estimation. Although a drawback of our top-down based approach is that the running time would increase as the persons in the scene increase, our parallel processing pipeline greatly redeem this deficiency. According to the statistics by OpenPose~\cite{cao2018openpose}, our library is more efficient than it when there are less than 20 persons in the scene. \red{From Fig.~\ref{fig:speedacc}(b) we can see that our pose tracking achieves on-par performance with the state-of-the-art library while running with high efficiency.}

\section{Conclusion}
In this paper, we propose a unified and realtime framework for multi-person fullbody pose estimation and tracking. To the best of our knowledge, it is the first framework that serves this purpose. Several novel techniques are presented to achieve this goal and we demonstrate superior performance in both efficacy and efficiency. A new dataset that contains full body keypoints (136 keypoints for each person) is annotated to facilitate the research in this area. We also present a standard library that is highly optimized for easy usage and hope that it can benefit our community. For our future research, we will also include 3D keypoints and mesh to our library.

\small{
\section*{ACKNOWLEDGMENT}
This work is supported in part by the National Key R\&D Program of China, No. 2017YFA0700800, Shanghai Municipal Science
and Technology Major Project (2021SHZDZX0102), Shanghai Qi Zhi Institute, SHEITC (2018-RGZN-02046). We appreciate Chenxi Wang for help developing the MXNet version and Yang Han for developing the Jittor version of AlphaPose. Hao-Shu Fang would like to thank the support from Baidu, MSRA and ByteDance Fellowship.
}


%





\ifCLASSOPTIONcaptionsoff
  \newpage
\fi



%
\bibliographystyle{IEEEtran}
\bibliography{egbib}





\end{document}